\PassOptionsToPackage{table,HTML}{xcolor}
\documentclass[sigplan,10pt]{acmart}

\usepackage{algorithm}
\usepackage{algpseudocode}
\usepackage{multirow}
\usepackage{placeins}
\usepackage{newunicodechar}

\usepackage[T1]{fontenc}

\newcommand{\wq}[1]{{\color{black} #1}}

\newcommand{\sys}{Mosaic} 

\AtBeginDocument{%
  }

\copyrightyear{2026}
\acmYear{2026}
\setcopyright{cc}
\setcctype{by}
\acmConference[SIGOPS ATC 2026]{The 2026 ACM SIGOPS Annual Technical Conference}{November 15--18, 2026}{Hong Kong, China}
\acmBooktitle{The 2026 ACM SIGOPS Annual Technical Conference (SIGOPS ATC 2026), November 15--18, 2026, Hong Kong, China}
\acmDOI{10.1145/3828523.3857033}
\acmISBN{979-8-4007-2818-1/2026/11}

\ccsdesc[500]{Computer systems organization~Architectures}
\ccsdesc[500]{Computing methodologies~Parallel computing methodologies}

\begin{document}

\title[Mosaic: Efficient CPU--GPU Coordinated Block-Sparse Attention]{Mosaic: Efficient CPU--GPU Coordinated Block-Sparse Attention for Long-Context Decoding}

\begin{abstract}
Long-context LLM serving increasingly relies on CPU-resident context KV cache, driven by KV offloading and disaggregated prefill--decode execution. Although block-sparse attention reduces KV access, existing designs remain inefficient:
GPU-centralized execution fetches selected KV blocks from CPU memory at each decoding step, making it PCIe-bound, while CPU--GPU partitioned execution places CPU-side sparse attention on the critical path, leaving substantial GPU time idle.
This paper presents Mosaic, a CPU--GPU coordinated block-sparse attention system for efficient long-context decoding with CPU-resident context KV. Mosaic builds on three key insights: output-contribution-aware KV budgeting, head-specific and granularity-aware sparse configuration, and cross-device coordinated scheduling for sparse tasks over CPU-resident KV. Guided by these insights, Mosaic integrates a lightweight head-property predictor, a granularity--budget selector, and a priority-based scheduler to automatically configure and coordinate sparse attention, preserving output quality while improving efficiency on heterogeneous architectures. 
\wq{Across 2 models, 3 benchmarks, and 40 tasks, Mosaic preserves near-FULL quality while achieving a better overall quality--efficiency tradeoff than the evaluated baselines. Compared with the best-quality fixed Quest configuration, Mosaic improves TPOT by up to 3.7$\times$. We further validate Mosaic across four hardware configurations and in a large-model multi-GPU deployment.}
\end{abstract}

\settopmatter{printfolios=false,printccs=true,printacmref=true}



\author{Feiyu Yao}
\affiliation{%
  \institution{Beijing University of Technology}
  \city{Beijing}
  \country{China}
}
\email{yaofeiyu10@emails.bjut.edu.cn}

\author{Zhixiong Niu}
\affiliation{%
  \institution{Microsoft Research Asia}
  \city{Beijing}
  \country{China}
}
\email{zhniu@microsoft.com}

\author{Xiaqing Li}
\affiliation{%
  \institution{Beijing Jiaotong University}
  \city{Beijing}
  \country{China}
}
\email{xqli1@bjtu.edu.cn}

\author{Yongqiang Xiong}
\affiliation{%
  \institution{Microsoft Research Asia}
  \city{Beijing}
  \country{China}
}
\email{yqx@microsoft.com}

\author{Juan Fang}
\affiliation{%
  \institution{Beijing University of Technology}
  \city{Beijing}
  \country{China}
}
\email{fangjuan@bjut.edu.cn}

\author{Qian Wang}
\authornote{Corresponding author: Qian Wang.}
\affiliation{%
  \institution{Beijing University of Technology}
  \city{Beijing}
  \country{China}
}
\email{wangqian2020@bjut.edu.cn}

\maketitle


\section{Introduction}
\label{intro}

Long-context inference is becoming an increasingly important workload in LLM serving~\citeN{liu2025comprehensive, jiang2024minference, lai2025flexprefill, xiao2024infllm}. As context length grows, the KV cache generated during prefill can quickly outgrow GPU memory capacity, making CPU offloading with hierarchical KV management increasingly common in practical systems~\citeN{chen2024magicpig, lee2024infinigen, zhang2025pqcache}. Meanwhile, CPU-resident KV also arises in disaggregated prefill--decode architectures, where prefill workers transfer the generated
KV cache to the host memory of decode workers for subsequent decoding~\citeN{qin2025mooncake, liu2025lmcache}. Consequently, long-context decoding often needs to access a large CPU-resident KV cache through the relatively slow CPU--GPU interconnect.

Block-sparse attention is a practical way to mitigate this bottleneck. It selects only a small set of query-relevant KV blocks at each decoding step, reducing both CPU--GPU data movement and attention computation. We focus on block-sparse attention because block- or chunk-level sparsity has become a common design choice in recent long-context inference systems, and its explicit knobs---block size and KV budget---make it well suited for system-level quality--efficiency tradeoffs.

Existing sparse attention systems with CPU-resident KV mainly fall into two execution designs. \textbf{GPU-centralized designs}~\citeN{chen2024magicpig, pan2025instattention, gao2024cost, zhang2025pqcache, lee2024infinigen} perform block-sparse attention on the GPU. They keep metadata in GPU memory for fast selection and fetch the selected KV blocks from CPU memory at every decoding step, making performance PCIe-bound and incurring extra GPU-memory overhead. \textbf{CPU--GPU partitioned designs}~\citeN{deng2026scaling, liuretrievalattention, chen2025ktransformers} instead perform sparse attention over the CPU-resident KV cache on the CPU and return only attention outputs to the GPU. This reduces selected-KV transfers and avoids GPU metadata overhead, but puts CPU-side sparse attention on the decoding critical path. 
Our analysis shows that, although CPU--GPU partitioned designs consistently outperform GPU-centralized designs, they still leave substantial GPU time idle during decoding, and the idle fraction increases with context length and batch size~(\S\ref{analysis}). The bottleneck lies in the CPU-side sparse branch,
which includes top-$k$ selection and attention over selected CPU-resident KV blocks. This motivates our goal: reducing these CPU-side overheads and increasing CPU--GPU overlap, thereby improving sparse-attention efficiency on heterogeneous architectures.

To address this problem, we present \sys{}, a CPU--GPU coordinated block-sparse attention system for efficient long-context decoding with CPU-resident context KV. \sys{} is built on three key insights. \textbf{First}, KV budget allocation should be output-contribution-aware rather than score-based (\S\ref{insight1}). Since high attention scores do not necessarily imply large output contributions, score-based budgeting can cause large output deviations. \sys{} therefore allocates KV budget by jointly considering attention-output deviation and each head's relative contribution, reducing unnecessary budget for low-impact heads. \textbf{Second}, sparse configuration should be both head-specific and granularity-aware (\S\ref{insight2}). Attention heads differ in whether they need CPU-side block-sparse attention. For heads that do, block size and KV budget are predictably coupled. \sys{} therefore skips unnecessary CPU-side sparse processing and jointly selects block size and KV budget only for heads that require it, balancing output quality and CPU-side cost. \textbf{Third}, sparse tasks over CPU-resident context KV have a narrow CPU--GPU end-to-end latency gap (\S\ref{insight3}). PCIe transfer largely offsets the GPU path's compute advantage for data-movement-bound sparse attention, while the GPU often waits for the CPU-side sparse branch. \sys{} therefore schedules selected sparse tasks across CPU and GPU to improve overlap and reduce attention completion time.

Based on the above insights, \sys{} unifies sparse configuration, budget allocation and cross-device coordinated execution through three key components. (1) \textit{A lightweight head-property predictor} estimates the key sparse properties of each head online (\S\ref{sec:predictor}). It determines whether a head can bypass CPU-side sparse attention; for heads that require retrieval, it predicts the budget--granularity relationship. The predictor is trained with offline labels derived from output-deviation profiling, enabling quality-preserving budget decisions. (2) \textit{A granularity--budget selector} uses these predictions to choose the top-$k$ selection granularity and KV budget for heads requiring CPU-side sparse attention~(\S\ref{sec:Granularity-Budget}). The selector accounts for GQA's shared-KV structure and minimizes data access and sparse-attention overhead while satisfying the output-deviation threshold. (3) \textit{A priority-based scheduler} coordinates sparse tasks over CPU-resident KV across two execution paths (\S\ref{sec:scheduler}). It assigns higher priority to tasks with larger estimated data-access volume and dynamically dispatches tasks from a shared queue to CPU and GPU execution paths, enabling execution overlap and reducing attention completion time. Together, these components allow \sys{} to preserve output quality while improving sparse-attention efficiency on heterogeneous architectures for long-context decoding.

\wq{We evaluate \sys{} against four fixed Quest configurations~\cite{tang2024quest}, StreamingLLM~\cite{xiao2023streamingllm}, and MagicPIG~\cite{chen2024magicpig} on Qwen2.5-7B-Instruct~\cite{yang2024qwen2} and Llama-3.1-8B-Instruct~\cite{grattafiori2024llama} across LongBench~\cite{bai2024longbench}, L-Eval~\cite{an-etal-2024-l}, and RULER~\cite{li-etal-2025-ruler}. 
\sys{} achieves the best overall quality--efficiency tradeoff, preserving near-FULL quality while improving TPOT by 1.9$\times$--3.4$\times$ on Qwen and 2.5$\times$--3.7$\times$ on Llama over the best-quality fixed Quest configuration evaluated. We further validate \sys{} across four hardware configurations with different CPU--GPU performance balances and in a large-model multi-GPU deployment using GLM-4.5-Air~\cite{zeng2025glm}, a 110B MoE model.

}

To the best of our knowledge, this paper makes the following contributions.
\begin{itemize}

    \item \sys{} is the first system to make KV budget allocation output-contribution-aware rather than score-based.
    
    \item \sys{} is the first system to identify top-$k$ selection granularity as a key optimization dimension for CPU-side sparse attention, and to expose its predictable coupling with KV budget.
    
    \item We design and implement \sys{}\footnote{Code is available at: \href{https://github.com/optiq-lab/Mosaic}{https://github.com/optiq-lab/Mosaic}.}. It combines three components to automatically configure and coordinate sparse attention, preserving near-FULL quality while substantially improving long-context decoding efficiency. 


\end{itemize}
\section{Background and Motivation}
\label{background}


This section first introduces block-sparse attention (\S\ref{Preliminaries}), then presents two representative execution designs with CPU-resident input-context KV (\S\ref{categories}), and finally analyzes their limitations to motivate our work (\S\ref{analysis}).

\begin{figure*}[t]
\begin{center}
\centerline{\includegraphics[width=0.95\linewidth]{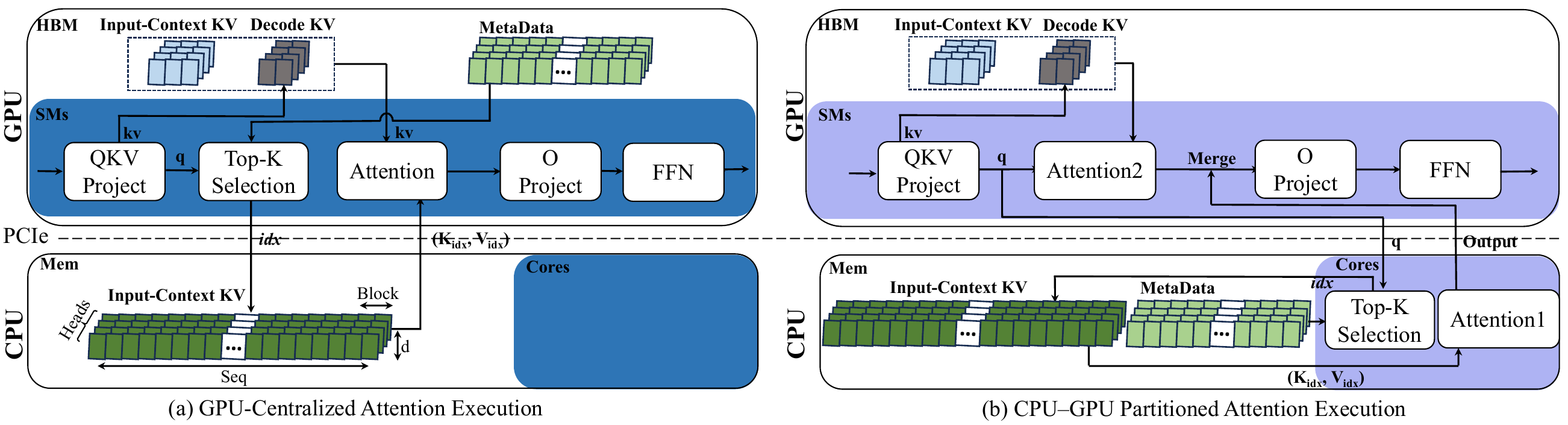}}
\vspace{-8pt}
\caption{Two execution designs for block-sparse attention with CPU-resident context KV.}
\vspace{-16pt}
\Description{}
\label{fig:attentionplacement}
\end{center}
\end{figure*}

\subsection{Block-Sparse Attention Preliminaries}
\label{Preliminaries}

Block-sparse attention reduces the cost of dense attention over the full KV cache through a two-stage procedure. The KV cache is partitioned into blocks, and each key block is represented by compact metadata. Given a query, the system first compares the query with the metadata of all key blocks to estimate the relevance of each block and then selects the top-$k$ blocks. Attention is subsequently computed only over the KV pairs associated with the selected blocks. 

A key property is that these blocks are logical rather than physical. Consequently, the logical block size can be adjusted dynamically without modifying the underlying storage layout. For instance, even if the smallest physical storage unit contains the keys for 16 tokens, the logical block size may still be configured as 16, 32, or 48 tokens. The logical block size determines the  granularity of block selection: larger logical blocks correspond to coarser-grained selection because each selected block spans more tokens.
Metadata for logical blocks can be constructed in different ways~\citeN{tang2024quest, chen2024arkvale, chen2025ktransformers, xiao2024infllm}. For example, Quest represents each block using the dimension-wise maxima and minima of its token keys~\cite{tang2024quest}. Although such metadata is much smaller than the KV cache, its GPU memory overhead remains nontrivial for large KV caches.

\subsection{Execution Designs of Block-Sparse Attention}
\label{categories}

\begin{figure*}[t]
\begin{center}
\centerline{\includegraphics[width=0.95\linewidth]{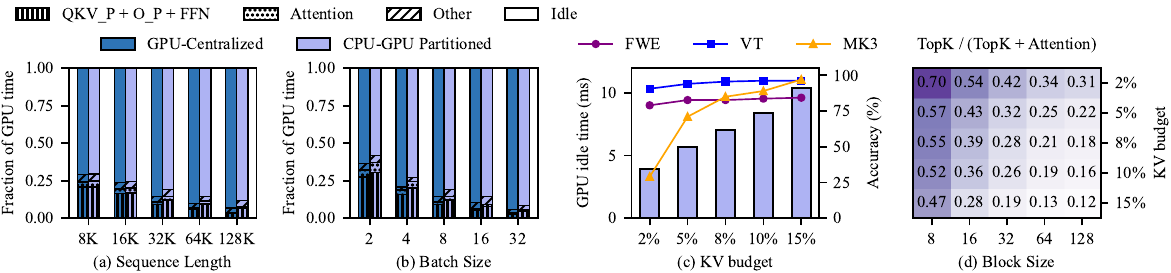}}
\vspace{-8pt}
\caption{Motivation for \sys{}. (a) Time breakdown across context lengths at batch size (BSZ)=4 and 5\% KV budget. (b) Time breakdown across batch sizes at SeqLen=32K and 5\% KV budget. (c) GPU idle time and accuracy across KV budgets at BSZ=8, SeqLen=32K. (d) CPU-side top-$k$ selection time fraction across block sizes and KV budgets.}
\vspace{-16pt}
\Description{}
\label{fig:motivation}
\end{center}
\end{figure*}

Given an input-context KV state already resident in CPU memory, existing sparse-attention systems primarily adopt two execution designs during decoding. Before describing these designs, we first clarify a common setting in which a small subset of KV is kept on the GPU, with dense attention computed over this subset to preserve local and critical context~\citeN{liuretrievalattention, chen2024magicpig,
xiao2024infllm, chen2025ktransformers, deng2026scaling}. This GPU-resident subset typically includes a small portion of the input-context KV and the newly generated decode KV. 
Under this common setting, the two execution designs differ in where sparse attention over the remaining CPU-resident KV is performed: either on the GPU or on the CPU, as shown in Figure~\ref{fig:attentionplacement}.



{\bf GPU--Centralized Attention Execution.} This class of methods~\citeN{zhou2025sparseserve, zhou2025progressive, huang2025nosa, takbir2025flexicache} centralizes sparse attention computation on the GPU, as shown in Figure~\ref{fig:attentionplacement}(a).
Key-block metadata for the CPU-resident context KV are maintained on the GPU to support fast top-$k$ selection. At each decoding step, the GPU computes block relevance scores by comparing the query vector against the metadata, selects the top-$k$ CPU-resident blocks, and retrieves the corresponding KV blocks from CPU memory using on-demand mechanisms, such as unified virtual addressing
(UVA). \wq{UVA allows GPU kernels to directly access CPU memory over PCIe without explicit CPU-side data movement.} The GPU then computes attention over both its resident KV subset and the fetched KV blocks. Since the selected blocks must be transferred from CPU memory over PCIe at every decoding step, this design remains constrained by PCIe bandwidth and incurs additional GPU memory overhead for metadata storage. 

{\bf CPU--GPU Partitioned Attention Execution.} This class of methods~\citeN{deng2026scaling, liuretrievalattention, chen2025ktransformers} partitions attention computation across the CPU and GPU, as shown in Figure~\ref{fig:attentionplacement}(b). The metadata for the CPU-resident KV portion are maintained in CPU memory. At each decoding step, the GPU computes dense attention over its resident KV subset, while the CPU uses the query vector received from the GPU to perform block-sparse attention over the CPU-resident KV. The CPU then returns its attention output to the GPU, where it is merged with the GPU dense-attention output. This design substantially reduces selected-KV transfers over PCIe, but places CPU-side block-sparse attention on the decoding critical path.

\subsection{Limitations of Existing Execution Designs}
\label{analysis}

We evaluate Quest~\cite{tang2024quest} under the two execution designs. Unless otherwise stated, we use a block size of 32 for top-$k$ selection. Following~\citeN{liuretrievalattention, chen2024magicpig,
xiao2024infllm, chen2025ktransformers}, we keep the initial 64 tokens and the 256 most recent tokens in GPU memory as the GPU-resident input-context KV subset.
We run Llama-3.1-8B-Instruct~\cite{grattafiori2024llama} on three RULER tasks---FWE, VT, and MK3---which represent aggregation, multi-hop tracing, and retrieval, respectively~\cite{li-etal-2025-ruler}. We use RULER because its configurable sequence lengths enable controlled evaluation across long-context settings.

Figures~\ref{fig:motivation}(a) and~\ref{fig:motivation}(b) show that both GPU-centralized and CPU--GPU partitioned attention execution leave a large fraction of GPU time idle. Even in the smallest setting, the GPU idle fraction remains above 60\%; it further increases with longer context lengths and larger batches, reaching nearly 90\% in the largest settings. Although CPU--GPU partitioned execution reduces the GPU idle time fraction compared with GPU-centralized execution, the remaining idle time fraction is still substantial. This indicates that statically assigning the sparse branch to the CPU does not fully utilize heterogeneous resources.

We next examine the source of this remaining idle time. In CPU--GPU partitioned execution, it mainly comes from the CPU-side block-sparse attention pipeline, which consists of top-$k$ block selection followed by attention over the selected blocks. This pipeline is affected by two key factors: the KV budget, which controls the amount of KV involved in selected-block attention, and the block granularity, which affects top-$k$ selection overhead. 

Reducing the KV budget naturally reduces attention computation. However, Figure~\ref{fig:motivation}(c) shows that uniformly reducing the KV budget across heads can quickly sacrifice quality. For example, reducing the budget from 15\% to 2\% lowers GPU idle time by 64.5\%, but drops MK3 accuracy from around 97\% to below 29\%. This suggests that simply applying the same budget reduction across all heads is not sufficient.
Block granularity is another important factor. As shown in Figure~\ref{fig:motivation}(d), at a 5\% KV budget, the fraction of top-$k$ selection time in \texttt{Top-K + Attention} decreases from 0.57 to 0.22 as the block size increases from 8 to 128. This shows that top-$k$ selection is a non-negligible and granularity-sensitive component of CPU-side overhead.

These observations motivate our goal: reducing the CPU-side sparse-branch overhead from top-$k$ selection and selected-block attention computation while preserving output quality, and improving CPU--GPU overlap by using otherwise idle GPU resources.

\section{Insights for Efficient CPU--GPU Sparse Attention}

This section presents three key insights that guide our design. All evaluations use Llama-3.1-8B-Instruct~\cite{grattafiori2024llama} on LongBench V2~\cite{bai2024longbench2}.

\subsection{Output-Contribution-Aware Budget Allocation}
\label{insight1}

We first clarify the relation between attention scores, attention outputs, and the final layer representation. For an attention head, the output is computed by scoring a query against keys, applying softmax normalization, and aggregating values with the resulting attention weights, as shown in Eq.~\ref{eq:attention}.
\begin{equation}
\label{eq:attention}
   o_{h} =
\mathrm{softmax}\!\left(\frac{q_{h} {K_{h}}^\top}{\sqrt{d}}\right)
V_{h}
\end{equation}

In multi-head attention, the outputs from all heads are concatenated and transformed by the output projection, as shown in Eq.~\ref{eq:output_projection}. 
\begin{equation}
\label{eq:output_projection}
    y = W_O [o_{1}; \ldots; o_{H}]
\end{equation}

Block-sparse attention uses block metadata to select a subset of KV blocks, and the calculation above is restricted to KV entries within the selected blocks. Thus, attention scores indicate which KV entries or blocks a head attends to, but may not fully capture how much the resulting head output affects the final representation after the output projection.

\subsubsection{Budgeting Should Target Attention Output, Not Attention Score}
\label{insight1-1}

\begin{figure}[t]
\begin{center}
\centerline{\includegraphics[width=0.90\linewidth]{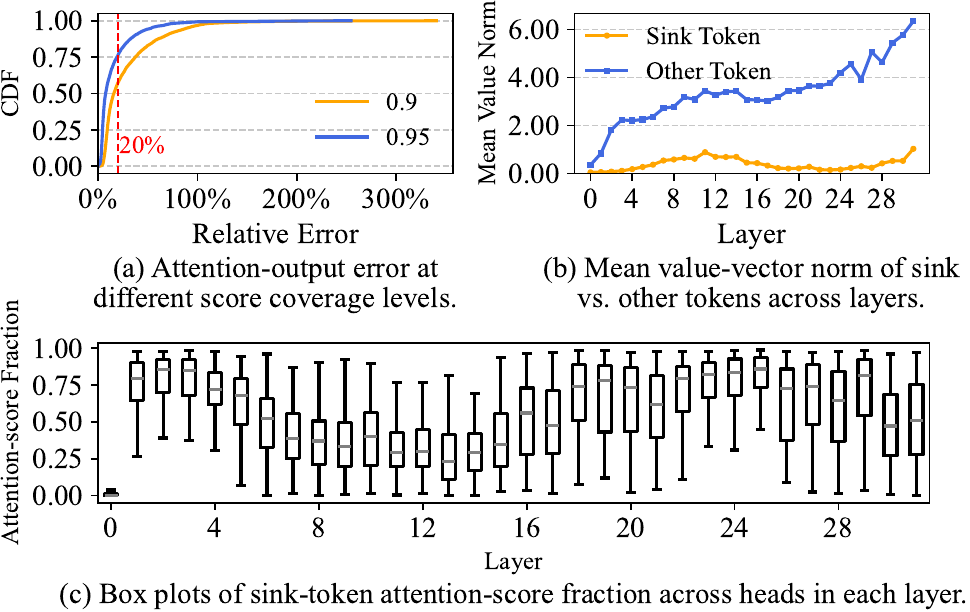}}
\vspace{-8pt}
\caption{Attention sink phenomenon breaks the link between attention-score coverage and output deviation.}
\vspace{-16pt}
\Description{}
\label{fig:Insight1-1}
\end{center}
\end{figure}

Existing sparse-attention methods~\citeN{zhang2025diffkv, zhou2025progressive, gao2025seerattention, zhu2025sampleattention, zhang2025spargeattention, xiao2024infllm, ribar2023sparq, singhania2024loki, liuretrievalattention, yuan2025native, lu2025moba} typically allocate KV budgets to preserve a target fraction of the total attention score, because attention scores provide a cheap proxy for token importance. However, the discussion above suggests a potential mismatch between score coverage and output fidelity. Figure~\ref{fig:Insight1-1}(a) confirms this mismatch: even when 95\% of the total attention score is preserved, 25\% of attention heads still incur a relative output error above 20\%.

A closer analysis suggests that this mismatch is related to the attention-sink phenomenon, where a small set of initial tokens consistently receives disproportionately large attention scores across layers and heads~\citeN{yang2025lserve, xiao2023streamingllm}.
As shown in Figure~\ref{fig:Insight1-1}(c), sinks account for a substantial fraction of the total attention score in most layers. Here, sinks refer to the initial 64 tokens, while other tokens denote all remaining tokens. However, Figure~\ref{fig:Insight1-1}(b) shows that sink tokens have small value norms, whereas other tokens exhibit much larger value norms that continue to grow in deeper layers. As a result, score-based budgeting can overestimate sink tokens and underestimate tokens with lower scores but larger output contributions. These observations suggest that budgeting should target attention-output deviation rather than attention-score coverage, by choosing the minimum budget that satisfies a target output-error threshold.

\begin{figure}[t]
\begin{center}
\centerline{\includegraphics[width=0.90\linewidth]{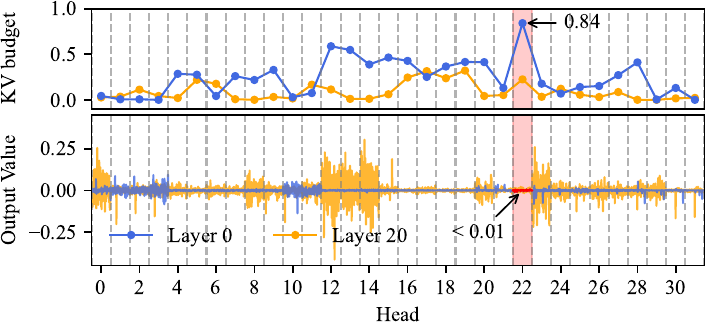}}
\vspace{-8pt}
\caption{Per-head output deviation alone is insufficient for KV budget allocation. Llama-3.1-8B-Instruct has 32 heads, each with 128 output dimensions.}
\vspace{-16pt}
\Description{}
\label{fig:Insight1-2}
\end{center}
\end{figure}

\subsubsection{Not All Heads Should Receive the Same Output-Deviation Budget}
\label{insight1-2}

Although budgeting by output deviation is closer to the final approximation target than attention-score coverage, it implicitly assumes that errors from different heads are equally important. In practice, head-output magnitudes vary substantially, so the same
relative output deviation may have different impacts across heads. 


Figure~\ref{fig:Insight1-2} illustrates this heterogeneity using two representative Transformer layers. The upper panel shows the minimum budget required for each head to keep its relative output deviation below 10\%, while the lower panel shows the corresponding full-attention output values for each dimension of each head. Some heads require relatively high budgets despite having much smaller output magnitudes than the dominant heads. Thus, enforcing the same relative output-deviation constraint for every head can still allocate substantial KV budgets to heads with relatively small output magnitudes, motivating us to further account for the relative contribution of different heads.


\wq{Since the final layer representation also depends on the output projection $W_O$ in Eq.~\ref{eq:output_projection}, we further examine whether the pre-$W_O$ head-output magnitude can serve as
a reliable proxy for post-projection contribution. We measure the post-projection contribution of head $h$ as $\|o_h W_h\|_2$, where $W_h$ denotes the submatrix of $W_O$ corresponding to head $h$. Across all layers of Llama-3.1-8B-Instruct, regressing $\log\|o_h W_h\|_2$ on $\log\|o_h\|_2$ achieves an average $R^2$ of 94.2\%, which increases only to 96.7\% when $\log\|W_h\|$ is additionally included. Therefore, the pre-$W_O$ head-output magnitude captures most of the variation in post-projection contribution and serves as a lightweight proxy for relative head contribution, avoiding the overhead of explicitly evaluating per-head output projection during online KV budget allocation.}

\textit{Implication: KV budget allocation should consider not only per-head output deviation but also each head's relative contribution, reducing unnecessary budget for low-impact heads while preserving output quality.}

\begin{figure}[t]
\begin{center}
\centerline{\includegraphics[width=0.98\linewidth]{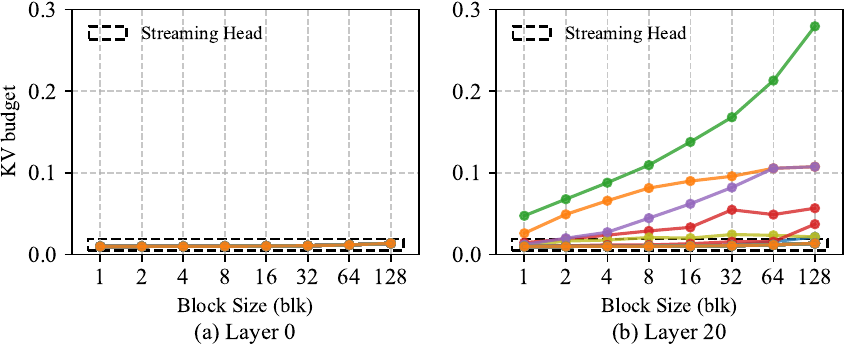}}
\vspace{-8pt}
\caption{Per-head minimum budget vs. block size for 32 heads in a layer. Each line represents one head.}
\vspace{-16pt}
\Description{}
\label{fig:Insight2}
\end{center}
\end{figure}

\subsection{Selection Granularity–Budget Coupling}
\label{insight2}

\S\ref{analysis} shows that CPU-side top-$k$ selection overhead is sensitive to block granularity. \wq{We denote the logical block size by $blk$.} A larger $blk$ reduces the number of candidate blocks and lowers selection overhead. However, block granularity cannot be tuned independently of the KV budget. Larger blocks make selection
coarser: each selected block covers more KV entries, which can increase approximation error at a fixed budget and thus require a higher KV budget to satisfy the same relative attention-output deviation threshold. To characterize this granularity--budget coupling, we analyze two representative Transformer layers and, for each head, quantify the minimum budget required at different block sizes to satisfy the relative attention-output deviation criterion introduced in \S\ref{insight1}. 


Figure~\ref{fig:Insight2} reveals clear heterogeneity across heads. One class of heads requires only a small and nearly constant budget across block sizes. For example, all heads in Layer~0 and most heads in Layer~20 exhibit this behavior. For these heads, attention over the GPU-resident KV entries is often sufficient to satisfy the target threshold, without CPU-side top-$k$ selection or sparse attention over CPU-resident KV. We refer to these heads as \textbf{streaming heads}. In contrast, for the remaining heads, the required budget increases systematically with block size; we refer to these as \textbf{retrieval heads}. Within the evaluated range, this increase is approximately linear in $log_2({blk})$. These results suggest that granularity--budget coupling only needs to be modeled for retrieval heads: streaming heads can bypass CPU-side top-$k$ selection and sparse attention, whereas retrieval heads require joint optimization of block size and budget to balance CPU-side overhead against approximation quality.

\textit{Implication: Streaming heads can bypass CPU-side top-$k$ selection and sparse attention, while retrieval heads require joint optimization of selection granularity and KV budget.}


\subsection{Narrow CPU--GPU Latency Gap}
\label{insight3}

For CPU-resident KV, decoding-time attention is fundamentally data-movement-bound rather than compute-bound~\cite{cao2025moe}. Consequently, even transferring only a small selected subset of KV to the GPU does not necessarily yield lower end-to-end latency than computing sparse attention directly on the CPU. The reason is that sparse attention reduces transferred bytes and attention FLOPs roughly in proportion, and therefore does not materially improve effective operational intensity relative to PCIe transfer. 

This, however, does not imply a large practical latency gap between CPU and GPU. Consider Llama-3.1-8B-Instruct under a 32K context and a 5\% KV budget. Each layer accesses only about 1.6K selected KV tokens, corresponding to roughly 6.4 MB per layer and 204.8 MB across 32 layers under BF16. If sparse attention is executed on the GPU, transferring these KV over a 32 GB/s PCIe link already takes about 6.4 ms. By contrast, executing sparse attention directly on the CPU requires only host-local memory access, which takes about 2 ms under a 100 GB/s memory bandwidth for the same data volume. More importantly, in CPU--GPU partitioned attention execution, the GPU is not continuously saturated, but instead remains idle while waiting for the CPU-side sparse branch, as shown in Figure~\ref{fig:motivation}(a) and \ref{fig:motivation}(b). 

\textit{Implication: Sparse attention should not be pinned to either device. Instead, it should be partitioned and overlapped across CPU and GPU to improve resource utilization and reduce overall attention completion time.}

\section{Design Overview}
\label{overview}


\begin{figure}[t]
\begin{center}
\centerline{\includegraphics[width=0.90\linewidth]{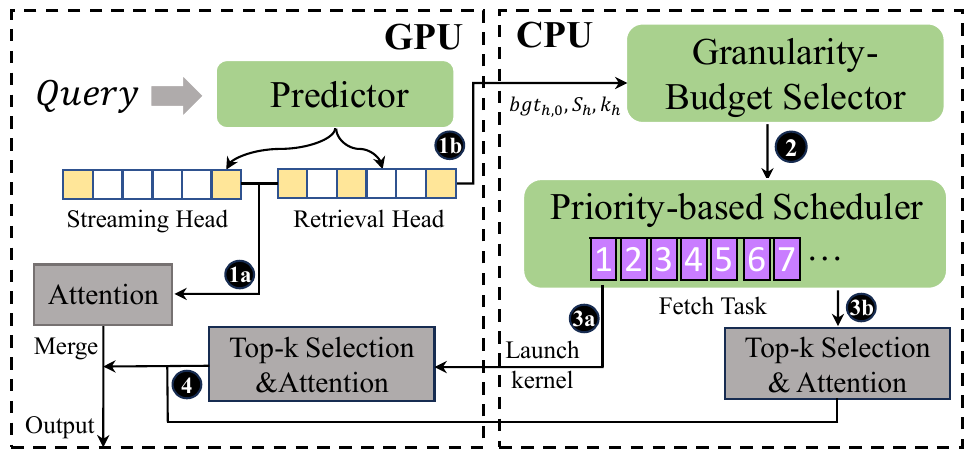}}
\vspace{-8pt}
\caption{The overview of \sys.}
\vspace{-16pt}
\Description{}
\label{fig:overview}
\end{center}
\end{figure}

To improve the efficiency of block-sparse attention on heterogeneous CPU-GPU architectures, we design \sys{}, an efficient CPU--GPU coordinated sparse-attention system for long-context
decoding. \sys{} consists of three key components. ($\mathrm{i}$) \textit{Head property predictor}: it identifies each attention head as either a streaming head or a retrieval head at low overhead and predicts the granularity-budget relationship for retrieval heads. ($\mathrm{ii}$) \textit{Granularity-budget selector}: it determines the optimal top-$k$ selection granularity and KV budget for each retrieval head, while remaining compatible with GQA. ($\mathrm{iii}$) \textit{Priority-based scheduler}: it dynamically dispatches sparse attention tasks according to task priority and the runtime status of the CPU and GPU.

Figure~\ref{fig:overview} presents the workflow of \sys{}. Given the query at each decoding step, \textcircled{1} \sys{} first uses the predictor to distinguish streaming and retrieval heads
and predict their head properties. \wq{For streaming heads, \textbf{(1a)} attention over the GPU-resident KV subset is sufficient, bypassing additional top-$k$ selection and attention over CPU-resident KV. For retrieval heads, \textbf{(1b)} the predicted head properties are sent to the selector.}
\textcircled{2} The selector determines the optimal top-$k$ selection granularity, KV budget, and a priority score for each retrieval head based on the predicted head properties. \textcircled{3} The  scheduler orders retrieval-head tasks by priority and places them into a unified priority queue.
\wq{For GPU execution, \textbf{(3a)} the scheduler selects a task from the priority queue and launches the corresponding sparse-attention kernel on the GPU. For CPU execution, \textbf{(3b)} CPU workers fetch tasks from the same queue and perform top-$k$ selection and attention over CPU-resident KV.} \textcircled{4} For each retrieval head, the sparse-attention output over CPU-resident KV is then fused with the dense-attention output over the GPU-resident KV subset to produce the final head
output.

\section{Head-Specific Sparse Configuration}











\begin{table*}[t]
\centering
\small
\setlength{\tabcolsep}{3pt}

\begin{tabular}{|c|c|}
\toprule
\textbf{Feature Group} &
\multicolumn{1}{c|}{\textbf{Features (number of features)}} \\
\midrule

Structural Features
& layer/head indices (2), CPU/GPU-resident KV lengths (2) \\

KV Distribution Features
& norm statistics of Key/Value vectors, including mean (6), variance (2), skewness (2), and kurtosis (2) \\

QK Interaction Features
& query--key interaction statistics for the anchor query (4) and current query (1)\\

Attention Contribution Features
& segment-wise log-sum-exp (6) and attention-output norm (5) \\

Query Dynamics Features
& query norms for the anchor/current queries (2) and their similarity (1) \\

Budget Estimation Features
& minimum KV budgets for the anchor query at four block granularities (4) \\

Cross-head Features
& maximum GPU-resident output norm across heads for the anchor/current queries (2) \\

\bottomrule
\end{tabular}
\caption{The 41-dimensional feature vector used by the predictor.}
\label{table:features}
\vspace{-16pt}
\end{table*}

\subsection{Challenge}

At each decoding step, the system must determine the type of each attention head and, for retrieval heads, estimate the granularity--budget relationship. These properties are query-dependent and vary across heads, layers, and decoding steps. Directly obtaining them online would require evaluating multiple sparse configurations per head, introducing prohibitive overhead. The challenge is to predict these dynamic head properties accurately enough for online sparse configuration while keeping the prediction overhead low.

\subsection{Offline Predictor Training}
\label{sec:predictor}

We first parameterize the relationship between budget and selection granularity for retrieval heads, then construct supervision signals for dynamic head properties through offline analysis, and finally train a lightweight predictor for online inference.

\textbf{Granularity-budget modeling.}
Our earlier analysis in \S\ref{insight2} shows that only \emph{retrieval heads} need CPU-side top-$k$ selection and sparse attention over CPU-resident KV, whereas \emph{streaming heads} can be served by the GPU-resident KV subset. Therefore, we model the budget behavior only for retrieval heads.
As shown in Figure~\ref{fig:Insight2}, the minimum KV budget required to satisfy a fixed output-error threshold increases with the logical block size ($blk$). Empirically, the relationship is approximately linear in the transformed $(log_2 (blk), bgt)$ space. For each retrieval head $h$, we thus fit a linear model in this transformed space using least-squares regression as 

\begin{equation}
\label{eq:bgt-blk}
{bgt}_h(blk) = bgt_{h,0} + k_h \cdot \log_2(blk),
\end{equation}
where $bgt_{h,0}$ is the minimum budget required at the finest granularity ($blk=1$), and $k_h$ captures how the required budget grows with the block size.

\textbf{Head property construction and labeling.} At each sampled decoding step, we abstract the behavior of each attention head $h$ into a set of predictable attributes, denoted as head properties $(bgt_{h,0}, k_h, s_h)$. Here, $bgt_{h,0}$ and $k_h$ characterize the budget–granularity relationship in Eq.~\ref{eq:bgt-blk}, while $s_h$ is a binary variable indicating whether the head is a streaming head. We model this as a supervised signal construction process: for each head at each sampled decoding step, we extract its optimal behavior through offline analysis and generate the corresponding labels.

Specifically, for a head $h$ at granularity $blk$, we define $bgt_h(blk)$ as the minimum budget satisfying the error constraint. Following \S\ref{insight1}, we define the error using attention-output deviation rather than attention scores, \wq{and normalize it by the maximum $L_2$ norm across all head outputs at the current decoding step, with the pre-$W_O$ head-output magnitude serving as a lightweight proxy for relative head contribution:}
\begin{equation}
\label{eq:min-bgt}
\frac{\left \|o_h(bgt, blk) - o_h\right \|_2}{\max_{h'=1}^{H} \left \|o_{h'}\right \|_2} \le \tau,
\end{equation}
where $\tau$ is the error threshold, whose impact is discussed in \S\ref{Ablation Study}; $H$ is the number of attention heads, $o_h$ is the full attention output of head $h$, and $o_h(bgt, blk)$ is the sparse attention output under budget $bgt$ and granularity $blk$.
\wq{Unlike a per-head relative-error constraint that normalizes each head by its own output magnitude, Eq.~\ref{eq:min-bgt} uses a shared normalization factor across heads. Consequently, heads with smaller relative contributions can tolerate larger errors relative to their own outputs and thus require smaller KV budgets, while heads with larger contributions are more strongly protected.
}

Based on Eq.~\ref{eq:min-bgt}, we construct training data through offline analysis. For each target model, we enumerate $(bgt, blk)$ configurations for all layers and heads at multiple decoding steps, with $blk \in \{1, 16, 32, 64, 128\}$, and compute the minimum budget satisfying $\tau$. Detailed sampling settings are provided in \S\ref{implementation}. We then fit the budget--granularity relationship to obtain the growth coefficient $k_h$, set $bgt_{h,0}$ to the minimum budget at $blk = 1$, and determine $s_h$ by comparing the output computed from only the initial and most recent KV against full attention: heads whose errors remain within $\tau$ are labeled as streaming heads; the others are labeled as retrieval heads. This yields the final property labels $(bgt_{h,0}, k_h, s_h)$ for each head at each sampled decoding step.

\textbf{Lightweight predictor.} We extract a 41-dimensional feature vector to characterize the state
of the current attention head, as summarized in Table~\ref{table:features}. \wq{CPU-resident KV statistics are computed and cached during prefill, while decoding updates query-dependent features using the cached statistics and GPU-resident KV, avoiding repeated access to CPU-resident KV.} The predictor is implemented as a feed-forward neural network with a shared backbone to jointly predict the head properties $(bgt_{h,0}, k_h, s_h)$. Specifically, $bgt_{h,0}$ and $k_h$ are modeled as continuous variables and optimized with regression objectives, while $s_h$ is a binary indicator of whether the head is streaming. The three prediction tasks share the same feature representation and are separated only at the output layer, improving feature reuse while keeping the model lightweight. \wq{The predictor contains 0.112M parameters and accounts for only 1.9\%--8.2\% of the decoding latency across our evaluated settings; a detailed analysis is in \S\ref{Ablation Study}.}

\subsection{Online Granularity and KV Budget Selection}
\label{sec:Granularity-Budget}

This module determines the top-$k$ selection granularity and KV budget for retrieval heads over the CPU-resident KV, with the goal of reducing metadata and KV access for bandwidth-bound attention.
We unify MHA and GQA by treating MHA as the special case of GQA with group size \(G=1\). For a GQA group with \(G\) heads sharing the same KV cache, all heads use a common selection granularity \(blk\), while each head may require a different KV budget under that granularity. We model the group-level data access cost as:
\begin{equation}
\label{eq:data-gqa}
T(blk)=2 \times \frac{L_{cpu}}{blk}+2 \times L_{cpu} \cdot \sum_{h=1}^{G} bgt_h(blk).
\end{equation}

Here, $L_{\mathrm{cpu}}$ denotes the number of keys stored in CPU memory. The first term captures metadata access for top-$k$ selection: following Quest, each key block is represented by its dimension-wise minimum and maximum values, giving the factor of 2. The second term captures KV access for sparse-attention computation, where the factor of 2 accounts for both keys and values. Since computing the exact union of KV entries selected by different heads is expensive, we use the sum of per-head budgets as an upper bound.

For each retrieval head $h$, $bgt_h(blk)$ denotes the minimum KV budget required to satisfy the error constraint under granularity $blk$, computed from Eq.~\ref{eq:bgt-blk} using the predicted head properties.
We enumerate all candidate granularities \(blk \in \{16,32,64,128\}\) and select the granularity \(blk^*\) that minimizes \(T(blk)\). Once the granularity is determined, the budget for each retrieval head is computed individually using Eq.~\ref{eq:bgt-blk}. This procedure is applied to any GQA group containing at least one retrieval head. 

\section{CPU-GPU Parallel Scheduling}
\label{Design3}

\begin{table}[t]
\centering
\footnotesize
\color{black}
\setlength{\tabcolsep}{4.5pt}
\begin{tabular}{|c|c|c|c|}
\toprule
\textbf{Platform} & \textbf{CPU (Physical cores)} & \textbf{GPU} & \textbf{PCIe} \\
\midrule
P1 & AMD EPYC 7V13 (96) & 4$\times$ A100 80GB & 4.0 $\times$16 \\
P2 & Xeon Gold 6348 (18) & A800 80GB & 4.0 $\times$16 \\
P3 & Xeon Platinum 8470Q (25) & RTX 5090 32GB & 5.0 $\times$16\\
P4 & Xeon Platinum 8470Q (22) & P6000 96GB & 5.0 $\times$16 \\
\bottomrule
\end{tabular}
\caption{\color{black}Hardware platforms used in our evaluation.}
\label{tab:testbeds}
\vspace{-16pt}
\end{table}

\subsection{Challenge}

Efficient CPU--GPU scheduling for sparse attention is challenging because task costs vary substantially across heads. Under GQA, multiple query heads share the same KV cache but may select different KV entries, making the selected-KV union size difficult to predict before execution. CPU-side sparse execution also exhibits high variance and long-tail latency due to irregular top-$k$ selection and attention. The challenge is therefore to schedule tasks efficiently across the CPU and GPU without relying on precise runtime prediction.

\subsection{Priority-Based Dynamic Scheduling}
\label{sec:scheduler}

To address this issue, we propose a priority-based dynamic scheduler. The key idea is to execute high-cost tasks earlier while allowing low-cost tasks to fill execution gaps. 

We define each scheduling unit as a task. For GQA, a task corresponds to one head group and includes top-$k$ selection over CPU-resident KV and the corresponding sparse attention computation. MHA is treated as a special case of GQA with group size $G=1$, where each task contains a single attention head. Groups whose heads are all streaming heads are excluded from scheduling, because they do not perform retrieval over CPU-resident KV. Only retrieval groups, i.e., groups containing at least one retrieval head,
are scheduled.
All tasks are generated at the granularity of a decoding batch and inserted into a shared priority queue before execution. We assign each task a priority based on the data access cost $T(blk)$ computed \wq{in Eq.~\ref{eq:data-gqa}}, with larger tasks receiving higher priority. At runtime, CPU workers dynamically fetch tasks from the shared priority queue, while the scheduler selects tasks from the same queue and launches the corresponding sparse-attention kernels on the GPU. Each CPU worker or GPU execution stream is assigned a new task only after completing its current task.

\section{Implementation}
\label{implementation}

\begin{table*}[t]
\centering

\fontsize{7.2pt}{8.2pt}\selectfont
\setlength{\tabcolsep}{3pt}

\begin{tabular}{|c|c|ccccccccccccc|c|}
\toprule

\textbf{Model}
& \textbf{Method}
& \textbf{FinQA}
& \textbf{GovRep}
& \textbf{Legal}
& \textbf{MeetSum}
& \textbf{MultiDoc}
& \textbf{NarrQA}
& \textbf{NatQA}
& \textbf{NewsSum}
& \textbf{Paper}
& \textbf{Patent}
& \textbf{RevSum}
& \textbf{SciQA}
& \textbf{TVShow}
& \textbf{Avg} \\

\midrule

\multirow{9}{*}{\rotatebox{90}{Qwen2.5-7B-Instruct}}

& FULL 
& 44.65 & 23.48 & 27.68 & 16.36 & 27.95 & 19.19 & 70.23 & 16.39 & 21.01 & 20.55 & 14.84 & 46.33 & 14.02 & 27.90 \\


& \cellcolor[HTML]{d4f0cf}{Quest(16,0.05)} & \cellcolor[HTML]{d4f0cf}{\textbf{45.82}} & \cellcolor[HTML]{d4f0cf}{21.99} & \cellcolor[HTML]{d4f0cf}{26.53} & \cellcolor[HTML]{d4f0cf}{15.80} & \cellcolor[HTML]{d4f0cf}{\textbf{28.09}} & \cellcolor[HTML]{d4f0cf}{18.91} & \cellcolor[HTML]{d4f0cf}{70.44} & \cellcolor[HTML]{d4f0cf}{15.61} & \cellcolor[HTML]{d4f0cf}{\textbf{20.73}} & \cellcolor[HTML]{d4f0cf}{\textbf{22.42}} & \cellcolor[HTML]{d4f0cf}{14.37} & \cellcolor[HTML]{d4f0cf}{45.12} & \cellcolor[HTML]{d4f0cf}{13.53} & \cellcolor[HTML]{d4f0cf}{27.64} \\

& Quest(16, 0.02)
& 42.75 & 22.42 & 26.34 & \textbf{16.17} & 26.26 & 19.02 & 69.61 & \textbf{15.87} & 20.39 & 20.46 & 13.76 & 41.96 & 11.71 & 26.67 \\

& Quest(32,0.05)
& 45.07 & 23.33 & 25.51 & 16.13 & 27.38 & 19.03 & 70.89 & 15.23 & 20.37 & 21.74 & 14.09 & 44.50 & 11.44 & 27.29 \\

& Quest(32,0.02)
& 42.28 & \textbf{23.34} & 24.88 & 15.80 & 25.80 & 17.96 & 68.15 & 15.24 & 19.89 & 22.00 & 14.03 & 39.79 & 12.90 & 26.31 \\

& StreamingLLM 
& 29.00 & 17.38 & 17.19 & 11.70 & 22.14 & 14.87 & 56.36 & 12.77 & 15.20 & 12.41 & 12.68 & 27.50 & 8.38 & 18.43 \\

& MagicPIG 
& 42.05 & 23.05 & 26.55 & 15.88 & 27.37 & 18.19 & 68.35 & 15.58 & 20.13 & 21.05 & \textbf{15.01} & 42.99 & 13.62 & 26.91 \\


& \cellcolor[HTML]{f2e6fa}{\sys} & \cellcolor[HTML]{f2e6fa}{44.59} & \cellcolor[HTML]{f2e6fa}{22.90} & \cellcolor[HTML]{f2e6fa}{\textbf{27.55}} & \cellcolor[HTML]{f2e6fa}{15.92} & \cellcolor[HTML]{f2e6fa}{27.38} & \cellcolor[HTML]{f2e6fa}{\textbf{19.62}} & \cellcolor[HTML]{f2e6fa}{\textbf{71.13}} & \cellcolor[HTML]{f2e6fa}{15.29} & \cellcolor[HTML]{f2e6fa}{20.46} & \cellcolor[HTML]{f2e6fa}{21.20} & \cellcolor[HTML]{f2e6fa}{14.49} & \cellcolor[HTML]{f2e6fa}{\textbf{45.32}} & \cellcolor[HTML]{f2e6fa}{\textbf{13.76}} & \cellcolor[HTML]{f2e6fa}{\textbf{27.66}} \\


& \cellcolor[HTML]{f2e6fa}{$\Delta$(FULL)} & \cellcolor[HTML]{f2e6fa}{-0.06} & \cellcolor[HTML]{f2e6fa}{-0.58} & \cellcolor[HTML]{f2e6fa}{-0.13} & \cellcolor[HTML]{f2e6fa}{-0.44} & \cellcolor[HTML]{f2e6fa}{-0.57} & \cellcolor[HTML]{f2e6fa}{+0.43} & \cellcolor[HTML]{f2e6fa}{+0.90} & \cellcolor[HTML]{f2e6fa}{-1.10} & \cellcolor[HTML]{f2e6fa}{-0.55} & \cellcolor[HTML]{f2e6fa}{+0.65} & \cellcolor[HTML]{f2e6fa}{-0.35} & \cellcolor[HTML]{f2e6fa}{-1.01} & \cellcolor[HTML]{f2e6fa}{-0.26} & \cellcolor[HTML]{f2e6fa}{-0.24} \\

\midrule

\multirow{9}{*}{\rotatebox{90}{Llama-3.1-8B-Instruct}}

& FULL
& 49.87 & 27.36 & 39.89 & 18.50 & 32.28 & 22.38 & 67.54 & 18.30 & 21.08 & 31.57 & 16.98 & 39.82 & 16.13 & 30.90 \\


& \cellcolor[HTML]{d4f0cf}{Quest(16,0.05)} & \cellcolor[HTML]{d4f0cf}{49.90} & \cellcolor[HTML]{d4f0cf}{26.89} & \cellcolor[HTML]{d4f0cf}{\textbf{39.54}} & \cellcolor[HTML]{d4f0cf}{18.34} & \cellcolor[HTML]{d4f0cf}{31.30} & \cellcolor[HTML]{d4f0cf}{23.40} & \cellcolor[HTML]{d4f0cf}{\textbf{67.55}} & \cellcolor[HTML]{d4f0cf}{\textbf{19.46}} & \cellcolor[HTML]{d4f0cf}{21.04} & \cellcolor[HTML]{d4f0cf}{29.68} & \cellcolor[HTML]{d4f0cf}{16.19} & \cellcolor[HTML]{d4f0cf}{39.32} & \cellcolor[HTML]{d4f0cf}{\textbf{16.07}} & \cellcolor[HTML]{d4f0cf}{\textbf{30.67}} \\

& Quest(16, 0.02)
& 49.14 & 25.86 & 38.25 & 18.12 & 30.54 & 22.72 & 66.78 & 18.41 & 20.67 & 27.81 & 15.94 & 39.48 & 15.12 & 29.91 \\

& Quest(32,0.05)
& 49.00 & 26.88 & 39.38 & 18.34 & 30.99 & 22.57 & 66.77 & 18.35 & 20.58 & 28.99 & 16.14 & \textbf{40.51} & 15.24 & 30.29 \\

& Quest(32,0.02)
& 45.64 & 24.82 & 36.10 & 17.53 & 28.06 & 22.87 & 64.59 & 16.86 & 20.18 & 24.53 & 16.10 & 37.83 & 15.42 & 28.50 \\

& StreamingLLM 
& 33.26 & 18.66 & 21.75 & 14.47 & 25.42 & 20.16 & 59.63 & 15.89 & 16.05 & 13.92 & 14.71 & 26.64 & 10.05 & 22.35 \\

& MagicPIG 
& 47.58 & 26.63 & 37.04 & 18.58 & 31.46 & 22.72 & 66.05 & 18.42 & \textbf{21.50} & 30.65 & \textbf{16.98} & 39.29 & 14.80 & 30.13 \\

 & \cellcolor[HTML]{f2e6fa} \sys
& \cellcolor[HTML]{f2e6fa} \textbf{50.05} & \cellcolor[HTML]{f2e6fa} \textbf{27.23} & \cellcolor[HTML]{f2e6fa} 39.08 & \cellcolor[HTML]{f2e6fa} \textbf{18.83} & \cellcolor[HTML]{f2e6fa} \textbf{31.82} & \cellcolor[HTML]{f2e6fa} \textbf{23.81} & \cellcolor[HTML]{f2e6fa} 67.26 & \cellcolor[HTML]{f2e6fa} 17.58 & \cellcolor[HTML]{f2e6fa} 21.15 & \cellcolor[HTML]{f2e6fa} \textbf{31.02} & \cellcolor[HTML]{f2e6fa} 16.20 & \cellcolor[HTML]{f2e6fa} 39.17 & \cellcolor[HTML]{f2e6fa} 15.14 & \cellcolor[HTML]{f2e6fa} 30.64 \\


& \cellcolor[HTML]{f2e6fa}{$\Delta$(FULL)} & \cellcolor[HTML]{f2e6fa}{+0.18} & \cellcolor[HTML]{f2e6fa}{-0.13} & \cellcolor[HTML]{f2e6fa}{-0.81} & \cellcolor[HTML]{f2e6fa}{+0.33} & \cellcolor[HTML]{f2e6fa}{-0.46} & \cellcolor[HTML]{f2e6fa}{+1.43} & \cellcolor[HTML]{f2e6fa}{-0.28} & \cellcolor[HTML]{f2e6fa}{-0.72} & \cellcolor[HTML]{f2e6fa}{+0.07} & \cellcolor[HTML]{f2e6fa}{-0.55} & \cellcolor[HTML]{f2e6fa}{-0.78} & \cellcolor[HTML]{f2e6fa}{-0.65} & \cellcolor[HTML]{f2e6fa}{-0.99} & \cellcolor[HTML]{f2e6fa}{-0.26} \\

\bottomrule
\end{tabular}

\caption{
Results on L-Eval Opened-ended tasks. $\Delta$ denotes the absolute score difference relative to FULL attention.
}
\label{tab:leval}
\vspace{-16pt}
\end{table*}

\begin{table*}[t]
\centering

\fontsize{7.2pt}{8.2pt}\selectfont
\setlength{\tabcolsep}{3pt}

\begin{tabular}{|c|c|cccccccccccccc|c|}
\toprule

\textbf{Model}
& \textbf{Method}
& \textbf{Qsp}
& \textbf{MFQA}
& \textbf{HotQA}
& \textbf{2WikiQA}
& \textbf{MNews}
& \textbf{TrivQA}
& \textbf{Sam}
& \textbf{PCnt}
& \textbf{PRetr}
& \textbf{Lcc}
& \textbf{MuSQ}
& \textbf{DReader}
& \textbf{QMS}
& \textbf{VCS}
& \textbf{Avg}
\\

\midrule

\multirow{9}{*}{\rotatebox{90}{Qwen2.5-7B-Instruct}}
& FULL
& 38.63 & 51.21 & 58.23 & 29.42 & 23.30 & 89.01 & 43.48
& 9.00 & 100.00 & 61.49 & 28.22 & 32.95 & 24.64 & 17.19 & 43.34 \\


& \cellcolor[HTML]{d4f0cf}{Quest(16,0.05)} & \cellcolor[HTML]{d4f0cf}{38.74} & \cellcolor[HTML]{d4f0cf}{53.31} & \cellcolor[HTML]{d4f0cf}{\textbf{56.98}} & \cellcolor[HTML]{d4f0cf}{29.67} & \cellcolor[HTML]{d4f0cf}{23.41} & \cellcolor[HTML]{d4f0cf}{\textbf{89.03}} & \cellcolor[HTML]{d4f0cf}{43.06} & \cellcolor[HTML]{d4f0cf}{9.00} & \cellcolor[HTML]{d4f0cf}{100.00} & \cellcolor[HTML]{d4f0cf}{60.68} & \cellcolor[HTML]{d4f0cf}{28.87} & \cellcolor[HTML]{d4f0cf}{32.03} & \cellcolor[HTML]{d4f0cf}{\textbf{24.20}} & \cellcolor[HTML]{d4f0cf}{16.33} & \cellcolor[HTML]{d4f0cf}{\textbf{43.24}} \\

& Quest(16,0.02)
& 39.60 & 52.95 & 55.58 & 29.26 & \textbf{23.42} & 87.39 & 44.00
& 9.00 & 100.00 & 60.08 & 27.96 & 32.54 & 23.47 & 16.11 & 42.95 \\

& Quest(32,0.05)
& 38.64 & 51.31 & 56.36 & 29.18 & 22.95 & 88.10 & 43.90
& 9.00 & 100.00 & 60.80 & 28.79 & 31.89 & 23.47 & 16.29 & 42.91 \\

& Quest(32,0.02)
& 38.19 & 51.32 & 53.83 & 28.23 & 22.89 & 88.33 & 42.93
& 9.00 & 100.00 & 57.01 & 27.31 & 30.08 & 23.42 & 15.41 & 42.00 \\

& StreamingLLM
& 20.46 & 22.89 & 39.50 & 21.76 & 17.91 & 82.70 & 40.74
& 9.00 & 16.00 & 52.02 & 17.62 & 22.70 & 18.68 & 13.42
& 26.81 \\

& MagicPIG
& 39.30 & 48.95 & 55.24 & \textbf{30.06} & 22.54 & 88.79 & \textbf{45.44}
& 9.00 & 95.00 & \textbf{61.10} & 29.39 & 29.77 & 23.26 & 16.60
& 42.46 \\


& \cellcolor[HTML]{f2e6fa}{\sys} & \cellcolor[HTML]{f2e6fa}{\textbf{39.60}} & \cellcolor[HTML]{f2e6fa}{\textbf{55.54}} & \cellcolor[HTML]{f2e6fa}{56.36} & \cellcolor[HTML]{f2e6fa}{29.05} & \cellcolor[HTML]{f2e6fa}{22.91} & \cellcolor[HTML]{f2e6fa}{87.24} & \cellcolor[HTML]{f2e6fa}{42.62} & \cellcolor[HTML]{f2e6fa}{\textbf{9.00}} & \cellcolor[HTML]{f2e6fa}{\textbf{100.00}} & \cellcolor[HTML]{f2e6fa}{59.46} & \cellcolor[HTML]{f2e6fa}{\textbf{29.78}} & \cellcolor[HTML]{f2e6fa}{\textbf{33.60}} & \cellcolor[HTML]{f2e6fa}{23.18} & \cellcolor[HTML]{f2e6fa}{\textbf{16.66}} & \cellcolor[HTML]{f2e6fa}{43.21} \\


& \cellcolor[HTML]{f2e6fa}{$\Delta$(FULL)} & \cellcolor[HTML]{f2e6fa}{+0.97} & \cellcolor[HTML]{f2e6fa}{+4.33} & \cellcolor[HTML]{f2e6fa}{-1.87} & \cellcolor[HTML]{f2e6fa}{-0.37} & \cellcolor[HTML]{f2e6fa}{-0.39} & \cellcolor[HTML]{f2e6fa}{-1.77} & \cellcolor[HTML]{f2e6fa}{-0.86} & \cellcolor[HTML]{f2e6fa}{+0.00} & \cellcolor[HTML]{f2e6fa}{+0.00} & \cellcolor[HTML]{f2e6fa}{-2.03} & \cellcolor[HTML]{f2e6fa}{+1.56} & \cellcolor[HTML]{f2e6fa}{+0.65} & \cellcolor[HTML]{f2e6fa}{-1.46} & \cellcolor[HTML]{f2e6fa}{-0.53} & \cellcolor[HTML]{f2e6fa}{-0.13} \\

\midrule

\multirow{9}{*}{\rotatebox{90}{Llama-3.1-8B-Instruct}}
& FULL
& 37.19 & 54.54 & 60.13 & 34.25 & 24.18 & 94.10 & 43.91
& 9.00 & 100.00 & 65.30 & 31.97 & 33.55 & 25.00 & 17.12 & 45.02 \\


& \cellcolor[HTML]{d4f0cf}{Quest(16,0.05)} & \cellcolor[HTML]{d4f0cf}{37.67} & \cellcolor[HTML]{d4f0cf}{55.15} & \cellcolor[HTML]{d4f0cf}{60.13} & \cellcolor[HTML]{d4f0cf}{33.95} & \cellcolor[HTML]{d4f0cf}{24.43} & \cellcolor[HTML]{d4f0cf}{94.10} & \cellcolor[HTML]{d4f0cf}{44.05} & \cellcolor[HTML]{d4f0cf}{9.00} & \cellcolor[HTML]{d4f0cf}{100.00} & \cellcolor[HTML]{d4f0cf}{64.84} & \cellcolor[HTML]{d4f0cf}{31.21} & \cellcolor[HTML]{d4f0cf}{31.70} & \cellcolor[HTML]{d4f0cf}{25.01} & \cellcolor[HTML]{d4f0cf}{16.22} & \cellcolor[HTML]{d4f0cf}{44.82} \\

& Quest(16,0.02)
& 38.49 & \textbf{57.06} & 60.13 & 33.97 & 23.87 & 94.10 & 43.33
& 10.00 & 99.00 & 65.15 & 30.76 & 32.10 & 24.73 & 16.02 & 44.91 \\

& Quest(32,0.05)
& \textbf{39.43} & 52.56 & 60.13 & 34.45 & \textbf{24.75} & 94.10 & \textbf{44.90}
& 9.33 & 100.00 & 65.52 & \textbf{31.69} & 31.54 & 25.21 & 16.21 & 44.99 \\

& Quest(32,0.02)
& 36.37 & 54.01 & 60.13 & 33.95 & 23.46 & 94.10 & 44.35
& 10.06 & 99.00 & \textbf{66.49} & 30.94 & 30.51 & 24.31 & 15.54 & 44.52 \\

& StreamingLLM
& 21.93 & 33.25 & 50.22 & 26.73 & 20.24 & 91.98 & 41.22
& 9.00 & 99.00 & 58.93 & 24.11 & 24.25 & 20.98 & 13.38 & 38.23 \\

& MagicPIG
& 36.10 & 53.15 & 58.31 & 33.66 & 24.28 & 93.81 & 44.34
& \textbf{13.12} & 99.00 & 62.83 & 31.62 & 32.74 & \textbf{25.37} & 16.42
& 44.63 \\


& \cellcolor[HTML]{f2e6fa}{\sys} & \cellcolor[HTML]{f2e6fa}{37.64} & \cellcolor[HTML]{f2e6fa}{56.93} & \cellcolor[HTML]{f2e6fa}{\textbf{60.13}} & \cellcolor[HTML]{f2e6fa}{\textbf{34.20}} & \cellcolor[HTML]{f2e6fa}{24.43} & \cellcolor[HTML]{f2e6fa}{\textbf{94.10}} & \cellcolor[HTML]{f2e6fa}{43.63} & \cellcolor[HTML]{f2e6fa}{9.00} & \cellcolor[HTML]{f2e6fa}{\textbf{100.00}} & \cellcolor[HTML]{f2e6fa}{64.95} & \cellcolor[HTML]{f2e6fa}{31.27} & \cellcolor[HTML]{f2e6fa}{\textbf{32.89}} & \cellcolor[HTML]{f2e6fa}{24.81} & \cellcolor[HTML]{f2e6fa}{\textbf{16.60}} & \cellcolor[HTML]{f2e6fa}{\textbf{45.04}} \\


& \cellcolor[HTML]{f2e6fa}{$\Delta$(FULL)} & \cellcolor[HTML]{f2e6fa}{+0.45} & \cellcolor[HTML]{f2e6fa}{+2.39} & \cellcolor[HTML]{f2e6fa}{+0.00} & \cellcolor[HTML]{f2e6fa}{-0.05} & \cellcolor[HTML]{f2e6fa}{+0.25} & \cellcolor[HTML]{f2e6fa}{+0.00} & \cellcolor[HTML]{f2e6fa}{-0.28} & \cellcolor[HTML]{f2e6fa}{+0.00} & \cellcolor[HTML]{f2e6fa}{+0.00} & \cellcolor[HTML]{f2e6fa}{-0.35} & \cellcolor[HTML]{f2e6fa}{-0.70} & \cellcolor[HTML]{f2e6fa}{-0.66} & \cellcolor[HTML]{f2e6fa}{-0.19} & \cellcolor[HTML]{f2e6fa}{-0.52} & \cellcolor[HTML]{f2e6fa}{+0.02} \\

\bottomrule
\end{tabular}

\caption{
LongBench quality results across sparse attention baselines. }
\label{tab:longbench}
\vspace{-16pt}
\end{table*}

\sys{} is implemented as an efficient block-sparse attention system, comprising approximately 2K lines of Python code and 3K lines of C++/CUDA code. We focus on optimizing the parallel execution of sparse attention, while all other operators are reused from the Hugging Face Transformers library~\cite{wolf-etal-2020-transformers} to ensure correctness and reproducibility.

\textbf{System Optimization.}
We implement a unified group-wise parallel attention framework for GQA across the CPU and GPU. On the CPU side, we accelerate BF16 computation using FBGEMM~\cite{fbgemm} and AVX2 vectorization, and parallelize attention using OpenMP~\cite{dagum1998openmp}. On the GPU side, we use multiple CUDA streams to reduce the overhead introduced by group-wise parallelism. We further implement optimized FlashDecoding~\cite{hong2023flashdecoding++} and top-$k$ kernels to accelerate inference. To reduce the data-movement overhead of accessing CPU-resident sparse KV blocks from the GPU, we leverage UVA
to enable direct access to CPU memory.

\textbf{Scheduling.}
To support efficient CPU--GPU coordinated execution, \wq{we dedicate one CPU core per GPU to scheduling and GPU kernel launches}, while a configurable set of CPU cores is assigned to sparse-attention computation. We ensure atomic task acquisition across CPU and GPU workers, as well as synchronization between kernel launch and execution. \wq{For multi-GPU execution, all devices are coordinated through the same priority queue: the scheduler dynamically selects sparse-attention tasks from the queue and launches them on available GPUs, without static task-to-GPU assignment.}

\textbf{Predictor.}
We train the predictor using LongBench V2~\cite{bai2024longbench2}, a real-world long-context benchmark. We randomly sample 50 examples with sequence lengths no shorter than 128K tokens for training, and use one additional non-overlapping example for testing. For each sequence, we process the context at 4K-token intervals and randomly sample 8 decoding steps at each interval to collect the 41-dimensional feature vector and its corresponding labels. For Llama-3.1-8B-Instruct, this sampling procedure produces about
12.5M training samples and 256K test samples. \wq{Generating the profiling labels takes approximately 8 hours per model, while training the 0.112M-parameter predictor takes only 2--4 minutes. Both are performed only once per model.} To avoid data leakage, we do not report LongBench V2 results in our evaluation.

\section{Evaluation}
\label{evaluation}

\subsection{Methodology}
\label{Setup}

\textbf{Testbed.} We evaluate \sys{} on four CPU--GPU platforms with different processor generations and interconnects, as summarized in Table~\ref{tab:testbeds}. Unless otherwise stated, experiments use P1 with one NVIDIA A100 80GB GPU and 24 physical CPU cores, 20 of which are dedicated to sparse-attention computation. \wq{P2--P4 are used for the hardware-sensitivity study. For the large-model multi-GPU evaluation, we use all four A100 GPUs and all 96 physical CPU cores on P1, with 40 CPU cores dedicated to sparse-attention computation.}

\noindent\textbf{Models.}
We evaluate Llama-3.1-8B-Instruct~\cite{grattafiori2024llama} and Qwen2.5-7B-Instruct~\cite{yang2024qwen2} in our main experiments, and \wq{additionally use GLM-4.5-Air~\cite{zeng2025glm}, a 110B MoE model, for the large-model multi-GPU evaluation.} The three models use GQA with 4, 8, and 12 query heads per KV group, respectively, and are evaluated with context lengths up to 128K. \wq{Model weights, activations, and KV caches use BF16.}

\noindent{\textbf{Workloads.}} We use two classes of workloads. \textbf{For generation quality}, we use L-Eval~\citeN{an-etal-2024-l} and LongBench~\citeN{bai2024longbench} for real-world long-context tasks, and RULER~\citeN{li-etal-2025-ruler} for controlled evaluation across context lengths. We use all 13 open-ended tasks in L-Eval and 14 LongBench tasks that do not overlap with L-Eval, retaining only LongBench samples longer than 8K tokens. For RULER, we use all 13 tasks, randomly sample 100 instances per task, and synthesize inputs with target lengths of 8K, 16K, 32K, 64K, and 128K tokens.  
\textbf{For efficiency evaluation}, we use RULER to precisely control input length and consider both fixed-length and mixed-length batches. In fixed-length batches, all requests have the same input length; in mixed-length batches, requests with randomly sampled context lengths are combined within the same batch.

\noindent\textbf{Baselines.}
We compare \sys{} with three representative sparse-attention methods: Quest~\cite{tang2024quest}, StreamingLLM~\cite{xiao2023streamingllm}, and MagicPIG~\cite{chen2024magicpig}. Quest performs block-sparse attention by selecting query-relevant KV blocks using block-level metadata. StreamingLLM uses window-based sparsity, retaining only attention sinks and a local window. MagicPIG uses locality-sensitive hashing (LSH) to sample query-relevant entries from CPU-resident KV and computes sparse attention on the CPU.

We use a common CPU-resident-KV setting, keeping 64 sink tokens and a 256-token local window on the GPU and the remaining historical KV on the CPU. Under this setting, Quest computes attention over retrieved KV blocks on the CPU; MagicPIG retains its original heterogeneous execution; and StreamingLLM attends only to GPU-resident KV.
\wq{We sweep Quest over different block sizes and KV budgets (\S\ref{sec:performance}) and report four representative configurations in the main comparisons}: Quest($blk$, $bgt$), where $blk\in\{16,32\}$ and $bgt\in\{0.02,0.05\}$ denote the block size and KV budget ratio.

\noindent{\textbf{Metrics.}} We evaluate generation quality and decoding efficiency. On RULER, we report question-answering accuracy. On LongBench and L-Eval, we report F1 and ROUGE. For efficiency, we report time per output token (TPOT), defined as the average time interval between consecutive output tokens during decoding, excluding the first output token. We first compute TPOT
for each request and then average it across all requests.

\begin{table}[t]
\centering

\scriptsize
\setlength{\tabcolsep}{3pt}

\begin{tabular}{|c|c|ccccc|c|}
\toprule

\textbf{Model} & \textbf{Method} & \textbf{8K} & \textbf{16K} & \textbf{32K} & \textbf{64K} & \textbf{128K} & \textbf{Avg} \\

\midrule
\multirow{9}{*}{\rotatebox{90}{Qwen2.5-7B-Instruct}}
& FULL     
& 92.52 & 92.34 & 88.42 & 71.34 & 25.74 & 74.07 \\

& \cellcolor[HTML]{d4f0cf}{Quest(16, 0.05)} & \cellcolor[HTML]{d4f0cf}{89.72} & \cellcolor[HTML]{d4f0cf}{90.24} & \cellcolor[HTML]{d4f0cf}{\textbf{88.21}} & \cellcolor[HTML]{d4f0cf}{69.23} & \cellcolor[HTML]{d4f0cf}{24.63} & \cellcolor[HTML]{d4f0cf}{72.41} \\

& Quest(16, 0.02)  
& 83.10 & 85.65 & 84.14 & 67.75 & 23.98 & 68.92 \\

& Quest(32, 0.05)  
& 86.23 & 87.35 & 84.99 & 69.40 & \textbf{25.26} & 70.65 \\

& Quests(32, 0.02)  
& 80.38 & 81.76 & 80.35 & 66.62 & 24.49 & 66.72 \\

& StreamingLLM 
& 18.17 &  16.82 & 12.66 & 9.60 & 6.75 & 12.80\\

& MagicPIG
& 66.20 & 67.27 & 60.23 & 40.04 & 15.14 & 49.78\\


& \cellcolor[HTML]{f2e6fa}{\sys} & \cellcolor[HTML]{f2e6fa}{\textbf{90.46}} & \cellcolor[HTML]{f2e6fa}{\textbf{90.39}} & \cellcolor[HTML]{f2e6fa}{87.28} & \cellcolor[HTML]{f2e6fa}{\textbf{69.57}} & \cellcolor[HTML]{f2e6fa}{24.94} & \cellcolor[HTML]{f2e6fa}{\textbf{72.53}} \\


& \cellcolor[HTML]{f2e6fa}{$\Delta$(FULL)} & \cellcolor[HTML]{f2e6fa}{-2.06} & \cellcolor[HTML]{f2e6fa}{-1.95} & \cellcolor[HTML]{f2e6fa}{-1.14} & \cellcolor[HTML]{f2e6fa}{-1.77} & \cellcolor[HTML]{f2e6fa}{-0.80} & \cellcolor[HTML]{f2e6fa}{-1.54} \\

\midrule
\multirow{9}{*}{\rotatebox{90}{Llama-3.1-8B-Instruct}}

& FULL     
& 94.29 & 90.65 & 85.93 & 84.68 & 73.52 & 85.81 \\

& \cellcolor[HTML]{d4f0cf}{Quest(16, 0.05)} & \cellcolor[HTML]{d4f0cf}{91.52} & \cellcolor[HTML]{d4f0cf}{89.78} & \cellcolor[HTML]{d4f0cf}{\textbf{85.66}} & \cellcolor[HTML]{d4f0cf}{\textbf{84.09}} & \cellcolor[HTML]{d4f0cf}{71.72} & \cellcolor[HTML]{d4f0cf}{84.55} \\

& Quest(16, 0.02)  
& 86.79 & 85.75 & 83.04 & 81.54 & 70.18 & 81.46 \\

& Quest(32, 0.05)  
& 88.35 & 86.58 & 83.26 & 81.93 & 70.63 & 82.15 \\

& Quest(32, 0.02)  
& 80.55 & 81.53 & 78.50 & 76.71 & 67.82 & 77.02 \\

& StreamingLLM 
& 16.23 & 16.19 & 13.34 & 13.41 & 9.79 & 13.79\\

& MagicPIG
& 89.93 & 87.78 & 84.30 & 82.40 & 68.76 & 82.63\\


& \cellcolor[HTML]{f2e6fa}{\sys} & \cellcolor[HTML]{f2e6fa}{\textbf{93.40}} & \cellcolor[HTML]{f2e6fa}{\textbf{90.64}} & \cellcolor[HTML]{f2e6fa}{85.03} & \cellcolor[HTML]{f2e6fa}{83.97} & \cellcolor[HTML]{f2e6fa}{\textbf{72.09}} & \cellcolor[HTML]{f2e6fa}{\textbf{85.03}} \\


& \cellcolor[HTML]{f2e6fa}{$\Delta$(FULL)} & \cellcolor[HTML]{f2e6fa}{-0.89} & \cellcolor[HTML]{f2e6fa}{-0.01} & \cellcolor[HTML]{f2e6fa}{-0.90} & \cellcolor[HTML]{f2e6fa}{-0.71} & \cellcolor[HTML]{f2e6fa}{-1.43} & \cellcolor[HTML]{f2e6fa}{-0.78} \\

\bottomrule
\end{tabular}
\caption{RULER quality across context lengths.}
\label{tab:ruler}
\vspace{-16pt}
\end{table}

\subsection{End-to-end Performance}
\label{sec:performance}

\textbf{Quality.} Tables~\ref{tab:leval} and~\ref{tab:longbench} report quality results on L-Eval and LongBench. Overall, \sys{} preserves generation quality well on realistic long-context workloads. On L-Eval, \sys{} trails FULL attention by only 0.24 and 0.26 on Qwen2.5-7B-Instruct and Llama-3.1-8B-Instruct, respectively, suggesting that open-ended generation is more sensitive to sparse approximation but still incurs limited degradation. On LongBench, the gap is negligible: \sys{} is 0.13 below FULL on Qwen2.5 and slightly exceeds FULL by 0.02 on Llama-3.1.

Compared with other baselines, \sys{} consistently achieves higher average quality than StreamingLLM and MagicPIG. Across the four model--benchmark pairs, StreamingLLM achieves only about 72\% of the
FULL-attention average score, whereas MagicPIG reaches about 98\% and \sys{} reaches 99.6\%. Among the evaluated fixed Quest configurations, \sys{} is only 0.03 lower than the best-quality configuration, Quest($16$, $0.05$).
Task-level results further show that the impact of sparsity varies across workloads. For QA tasks, \sys{} improves on NarrQA (+0.43/+1.43) and MFQA (+4.33/+2.39), but degrades on SciQA (-1.01/-0.65) and HotQA (-1.87/0.0). Quality degradation is also observed on some summarization, open-ended generation, and code tasks, including NewsSum (-1.10/-0.72), TVShow (-0.26/-0.99), and LCC (-2.03/-0.35).

Table~\ref{tab:ruler} further evaluates accuracy across context lengths on RULER. \sys{} achieves 72.53 and 85.03 on Qwen2.5-7B-Instruct and Llama-3.1-8B-Instruct, respectively, remaining within 1.54 and 0.78 points of FULL while slightly outperforming Quest($16$, $0.05$) on both models. StreamingLLM
suffers severe quality degradation, reaching only about 16--17\% of FULL-attention accuracy on average. MagicPIG is competitive on Llama-3.1-8B-Instruct but drops to 67.2\% of FULL accuracy on
Qwen2.5-7B-Instruct, which we attribute to outliers in several Q/K heads that reduce the effectiveness of LSH-based sampling. Overall, \sys{} maintains high quality across diverse tasks and
context lengths by adapting both top-$k$ selection granularity and KV budget allocation rather than relying on a single fixed sparse configuration.

\begin{figure}[t]
\begin{center}
\centerline{\includegraphics[width=0.6\linewidth]{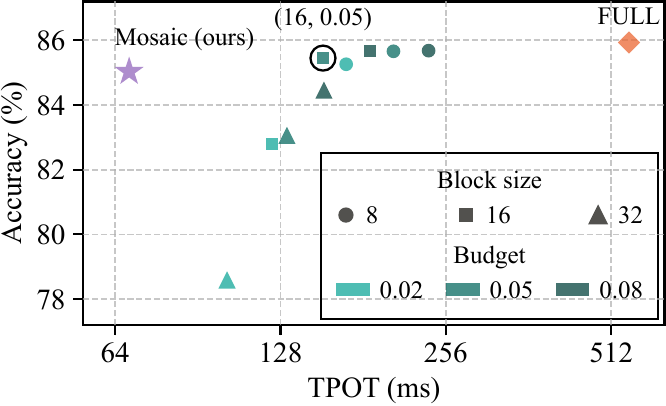}}
\vspace{-8pt}
\caption{\wq{Quality--efficiency tradeoff under different Quest configurations on RULER with Llama-3.1-8B-Instruct at 32K context length and batch size 4.}}
\vspace{-16pt}
\Description{}
\label{fig:quest-config}
\end{center}
\end{figure}

\begin{figure*}[t]
\begin{center}
\centerline{\includegraphics[width=0.90\linewidth]{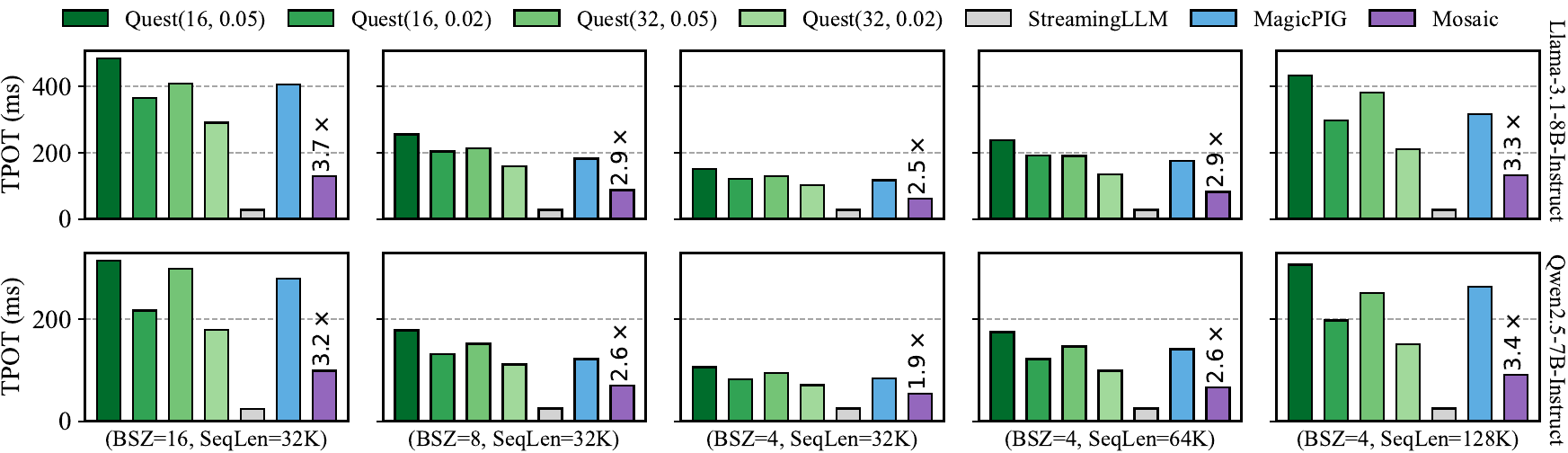}}
\vspace{-8pt}
\caption{TPOT comparison under different batch sizes and sequence lengths on RULER. The annotated speedups are relative to Quest($16$, $0.05$).}
\Description{}
\label{TBT}
\vspace{-16pt}
\end{center}
\end{figure*}

\begin{figure}[t]
\begin{center}
\centerline{\includegraphics[width=0.90\linewidth]{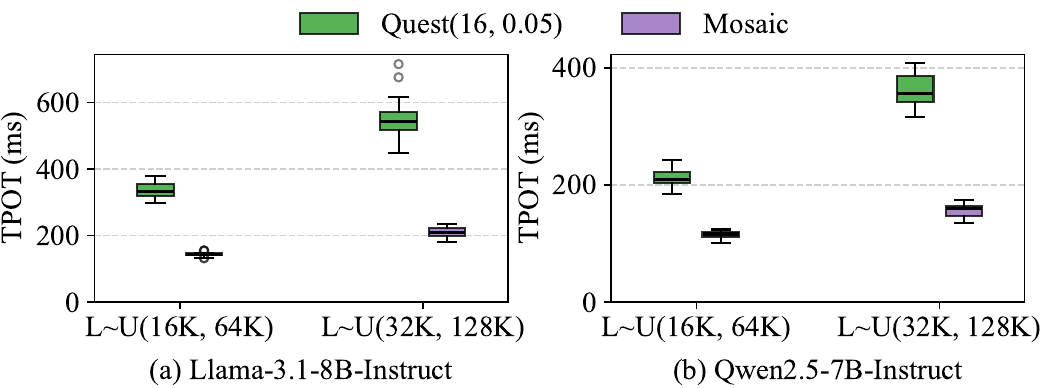}}
\vspace{-8pt}
\caption{TPOT under mixed-length workloads on RULER.}
\vspace{-16pt}
\Description{}
\label{fig:variedseq}
\end{center}
\end{figure}

\wq{
\noindent\textbf{Quality--efficiency tradeoff.} Figure~\ref{fig:quest-config} shows the quality--efficiency tradeoff of Quest under different block sizes and KV budgets, with \sys{} and FULL shown as references. The evaluated Quest configurations span a broad operating range, from efficiency-oriented settings with lower TPOT to more conservative settings that better preserve generation
quality. In particular, Quest(32, 0.02) already achieves low TPOT but incurs substantial quality degradation, suggesting that more aggressive configurations would provide limited additional value for quality-preserving inference. Among the four Quest configurations used in our main comparisons, Quest(16, 0.05) best preserves near-FULL
quality across workloads and context lengths. We therefore use it as the primary Quest reference in the following efficiency evaluation.
}

\textbf{Efficiency.} Figure~\ref{TBT} compares the TPOT of \sys{} with all baselines. The annotated speedups are normalized to Quest($16$, $0.05$), the fixed Quest configuration with the best quality. \sys{} achieves 2.5$\times$--3.7$\times$ speedup on Llama and 1.9$\times$--3.4$\times$ on Qwen. The gain increases with batch size or context length. For example, at batch size 4, the speedup on Qwen increases from 1.9$\times$ at 32K to 3.4$\times$ at 128K. 

Among fixed Quest configurations, Quest(32,0.02) is typically the fastest, but it is still slower than \sys{}. This gain comes from jointly adapting the granularity and KV budget, improving GPU utilization, and skipping top-$k$ selection and sparse attention for streaming heads.
StreamingLLM achieves the lowest TPOT, but only by discarding most contextual information, which leads to severe quality loss as shown in Table~\ref{tab:ruler}. MagicPIG preserves quality better than StreamingLLM, but remains 1.9$\times$--3.1$\times$ slower than \sys{} due to the overhead of hash-based sampling and CPU-side sparse attention. Overall, \sys{} maintains near-FULL quality while achieving lower TPOT than quality-preserving sparse baselines.

We further evaluate \sys{} against Quest(16,0.05) under mixed-length workloads over 20 runs, as shown in Figure~\ref{fig:variedseq}. In each run, we sample 8 sequences
uniformly from 16K to 64K and another 8 from 32K to 128K.
On both Llama and Qwen, \sys{} consistently achieves lower TPOT than Quest(16,0.05), and the gain becomes larger as the input-length range shifts toward longer contexts. Specifically, on Llama, the median TPOT speedup increases from about 2.3$\times$ under $L\sim U(16K,64K)$ to 2.6$\times$ under $L\sim U(32K,128K)$. This suggests that \sys{} becomes increasingly beneficial as attention accounts for a larger fraction of decoding time in longer contexts.

Table~\ref{tab:Idle} reports the average GPU idle time at each attention layer during decoding, together with its ratio to the total execution time. Overall, \sys{} significantly reduces GPU idle time ratio compared with Quest(16,0.05). For example, under (4, 32K), the idle time on Qwen decreases from 2.33 ms to 0.78 ms, while the idle ratio drops from 70.65\% to 45.78\%.
As the context length increases, \sys{} eliminates increasingly more GPU idle time in absolute terms. At batch size 4, the reduction on Qwen increases from 1.55 ms at 32K to 3.32 ms at 64K and 6.17 ms
at 128K. However, the reduction in idle ratio becomes smaller at 128K, indicating that a larger fraction of the CPU--GPU imbalance remains exposed. This is because longer sequences place greater
pressure on CPU and PCIe bandwidth, making the resulting stalls more difficult to hide through CPU--GPU parallel execution.

\begin{table}[t]
\centering

\footnotesize
\setlength{\tabcolsep}{2pt}

\begin{tabular}{|c|cc|cc|}
\toprule
\multirow{2}{*}{\textbf{\shortstack{(BSZ,\\SeqLen)}}} &
\multicolumn{2}{c|}{\textbf{Qwen2.5-7B-Instruct}} &
\multicolumn{2}{c|}{\textbf{Llama-3.1-8B-Instruct}} \\
 & \textbf{Quest(16, 0.05)} & \textbf{\sys} & \textbf{Quest(16, 0.05)} & \textbf{\sys} \\
 
\midrule
(16, 32K) & 8.72 / 88.79\% & 2.16 / 69.64\% & 13.23 / 87.42\% & 3.00 / 74.36\% \\
(8, 32K)  & 4.54 / 81.46\% & 1.23 / 56.27\% & 6.91 / 86.45\% & 1.69 / 61.79\% \\
(4, 32K)  & 2.33 / 70.65\% & 0.78 / 45.78\% & 4.06 / 85.83\% & 1.00 / 52.16\% \\
(4, 64K)  & 4.43 / 81.30\% & 1.11 / 53.68\% & 6.43 / 86.50\% & 1.47 / 57.50\% \\
(4, 128K) & 8.24 / 86.17\% & 2.07 / 73.14\% & 11.61 / 85.89\% & 3.12 / 75.45\% \\
\bottomrule
\end{tabular}
\caption{Average GPU idle time and idle ratio (ms / \%) at each attention layer during decoding on RULER.}
\label{tab:Idle}
\vspace{-16pt}
\end{table}

\begin{figure}[t]
\begin{center}
\centerline{\includegraphics[width=0.90\linewidth]{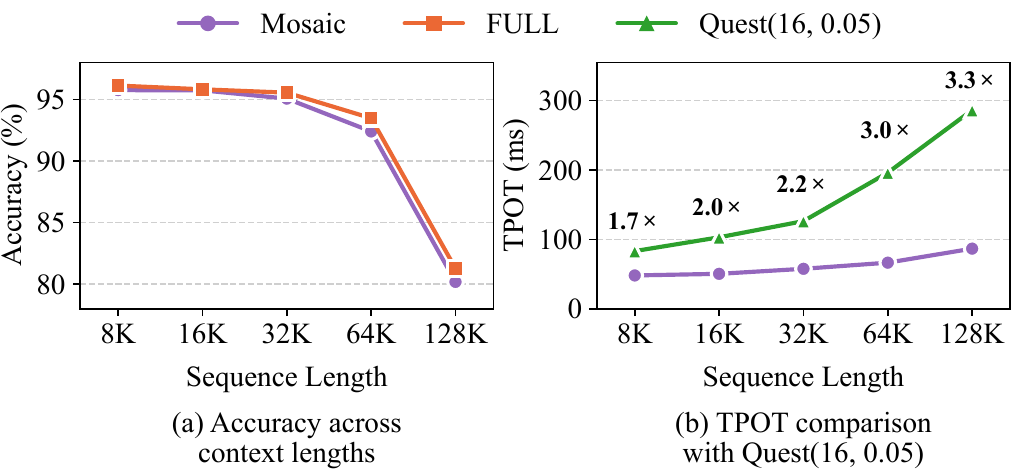}}
\vspace{-8pt}
\caption{\color{black}Quality and efficiency of \sys{} on GLM-4.5-Air with four A100 GPUs.}
\vspace{-16pt}
\Description{}
\label{fig:glm4.5}
\end{center}
\end{figure}

\wq{
\subsection{Generality across Deployment Settings}

We evaluate \sys{} across different hardware configurations and a large-model multi-GPU deployment on RULER with a batch size of 4. 

\textbf{Across Hardware Platforms.} We evaluate \sys{} on four platforms (P1--P4) with different CPU, GPU, and PCIe configurations, as detailed in Table~\ref{tab:testbeds}. As shown in Table~\ref{tab:cross-platforms}, CPU-side sparse attention provides higher effective bandwidth on P1 and P2, whereas GPU-side execution provides higher effective bandwidth on P3 and P4. Despite this reversed CPU--GPU performance balance, \sys{} consistently
achieves 1.48$\times$--1.71$\times$ speedup over the faster of CPU-only and GPU-only execution, approaching the ideal speedups of 1.65$\times$--1.95$\times$ under perfect CPU--GPU load balancing. These results show that \sys{} remains effective across the evaluated hardware configurations without relying on a particular CPU--GPU performance balance.

\textbf{A Large-Model Multi-GPU Case.} We further evaluate \sys{} on GLM-4.5-Air, a 110B MoE model, using expert parallelism for its MoE layers. As shown in Figure~\ref{fig:glm4.5}, \sys{} maintains RULER accuracy
close to FULL across context lengths from 8K to 128K. At 128K, \sys{} achieves an accuracy of 80.18, compared with 81.27 for FULL. Meanwhile, \sys{} reduces TPOT compared with Quest(16, 0.05), achieving 1.7$\times$--3.3$\times$ speedup from 8K to 128K. This experiment demonstrates the effectiveness of \sys{} in this large-model multi-GPU setting, rather than scalability across different GPU counts or parallelization strategies. 
}

\begin{table}[t]
\centering
\color{black}
\footnotesize
\setlength{\tabcolsep}{3pt}
\begin{tabular}{|c|c|c|c|c|}
\toprule
\multirow{2}{*}{\textbf{Platform}} 
& \multicolumn{2}{c|}{\textbf{Effective bandwidth (GB/s)}}
& \multicolumn{2}{c|}{\textbf{Speedup}} \\
\cmidrule(lr){2-3} \cmidrule(lr){4-5}
& \textbf{CPU} & \textbf{GPU (PCIe)} & \textbf{Ideal} & \textbf{Measured} \\
\midrule
P1 & 37.60 & 24.40 & 1.65$\times$ & 1.50$\times$ \\
P2 & 30.97 & 22.92 & 1.74$\times$ & 1.48$\times$ \\
P3 & 43.51 & 48.80 & 1.89$\times$ & 1.70$\times$ \\
P4 & 45.98 & 48.34 & 1.95$\times$ & 1.71$\times$ \\
\bottomrule
\end{tabular}
\caption{\color{black}Performance of \sys{} across different hardware platforms at 32K context length.
Speedup is relative to the faster of CPU-only and GPU-only execution.}
\label{tab:cross-platforms}
\vspace{-16pt}
\end{table}

\subsection{Ablation Study}
\label{Ablation Study}

All ablation studies in this section are conducted on RULER with Llama-3.1-8B-Instruct. Unless the experiment involves mixed-length workloads, we use an input length of 32K and a batch size of 4.

\begin{figure}[t]
\begin{center}
\centerline{\includegraphics[width=0.98\linewidth]{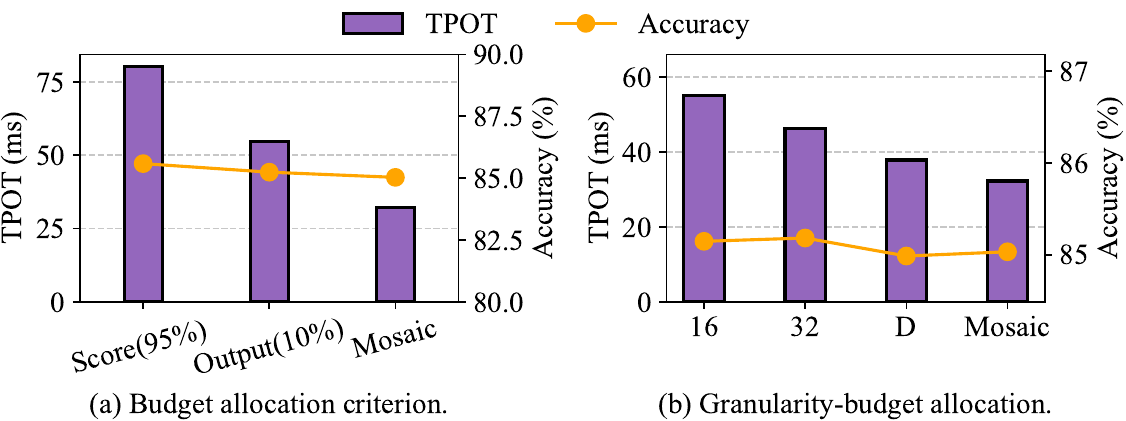}}
\vspace{-8pt}
\caption{Break down by components.}
\vspace{-16pt}
\Description{}
\label{fig:ablation}
\end{center}
\end{figure}

\textbf{KV budget allocation criterion.} We compare three KV budget allocation criteria. Score(95\%) allocates budget based on attention scores and retains enough KV to cover 95\% of the attention mass. Output(10\%) allocates the budget based on per-head relative output deviation, normalizing each
head by its own output magnitude, with the deviation threshold set to 10\%. \sys{} instead accounts for each head's relative contribution under the same 10\% output-deviation threshold, allowing the error tolerance to vary across heads. Figure~\ref{fig:ablation}(a) shows the resulting accuracy–latency tradeoff. Overall, \sys{} achieves the best tradeoff: it maintains accuracy close to Output(10\%), while further reducing TPOT from 54.60 ms to 32.28 ms; compared with Score(95\%), the latency reduction is even larger. 
These results show that attention-score-based allocation and uniform per-head relative-error constraints can over-allocate KV budgets. By accounting for relative head contribution, \sys{} allocates KV budgets more effectively, substantially reducing latency with little accuracy loss.

\textbf{Dynamic granularity--budget allocation and streaming head skipping.} Figure~\ref{fig:ablation}(b) compares four settings: fixed selection granularity 16 and 32, our proposed dynamic granularity--budget allocation (\textbf{D}) and the full method with additional streaming-head skipping (\sys{}). Both proposed designs provide clear incremental benefits. Compared with fixed granularity, \textbf{D} significantly reduces TPOT with almost no accuracy loss, showing the benefit of jointly adapting granularity and KV budget across heads. Adding streaming-head skipping further reduces TPOT from 38 ms to 32 ms while keeping accuracy nearly unchanged, showing that skipping top-$k$ selection and sparse-attention computation for streaming heads avoids unnecessary overhead. Overall, \sys{}
achieves the best accuracy--latency tradeoff.

\begin{table}[t]
\centering
\small
\setlength{\tabcolsep}{2pt}
\begin{tabular}{|c|cc|}
\toprule
\textbf{Scheduling Strategy}
& \textbf{$L \sim U(16K,64K)$}
& \textbf{$L \sim U(32K,128K)$} \\
\midrule
No parallel             & 218.56 & 312.08 \\
Uniform allocation      & 193.52 & 274.16 \\
Length-based allocation & 170.32 & 237.28 \\
Mosaic                  & 145.20 & 192.72 \\
\bottomrule
\end{tabular}
\caption{TPOT (ms) comparison of scheduling strategies.}
\label{tab:priority_scheduler}
\vspace{-16pt}
\end{table}

\textbf{Priority-based scheduler.} We compare four scheduling strategies on two mixed-length workloads, using the same sampling procedure as in the efficiency evaluation (\S\ref{sec:performance}). Specifically, we consider \textit{No Parallel}, which does not perform CPU--GPU parallel execution; \textit{Uniform allocation}, which uniformly distributes tasks according to the number of attention heads; \textit{Length-based allocation}, which statically assigns tasks based on input length; our priority-based scheduler. Table~\ref{tab:priority_scheduler} shows that \sys{} achieves the lowest TPOT under both workloads. Although Length-based allocation outperforms No Parallel and Uniform allocation, it still underperforms \sys{}. These results show that static length-based assignment cannot fully capture runtime task heterogeneity. In contrast, \sys{} prioritizes tasks by estimated cost and lets CPU and GPU workers dynamically retrieve them, yielding larger gains for longer-context workloads.

\begin{figure}[t]
\begin{center}
\centerline{\includegraphics[width=0.98\linewidth]{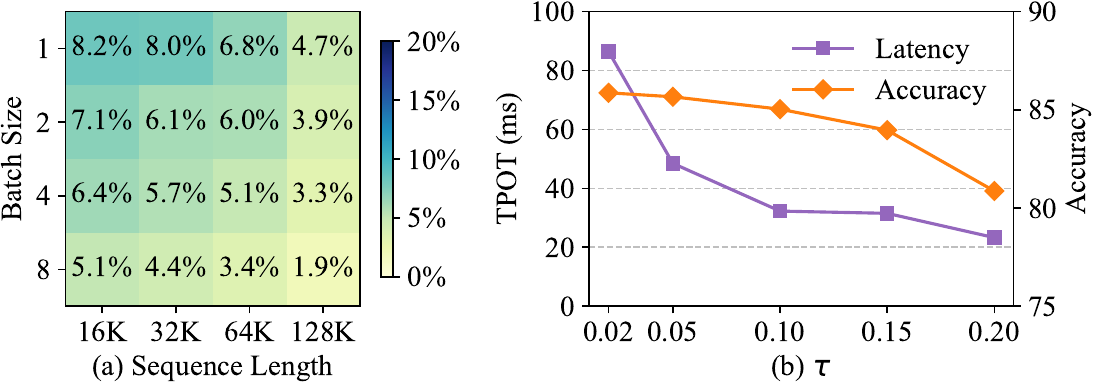}}
\vspace{-8pt}
\caption{(a) Predictor overhead. (b) Effect of $\tau$.}
\vspace{-16pt}
\Description{}
\label{fig:predictor-tau}
\end{center}
\end{figure}

\textbf{Predictor accuracy and overhead.} For Llama and Qwen, the predictor achieves $R^2$ values of 0.797/0.824 for $bgt_{h,0}$ and 0.663/0.557 for $k_h$, respectively, and F1 scores of 0.945/0.921 for streaming-head classification. The predictors are trained only on LongBench V2 and directly applied to LongBench, L-Eval, and RULER without workload-specific profiling or retraining. Together with the quality results in \S~\ref{sec:performance}, these findings suggest that the predictor provides useful estimates for \sys{}'s online configuration decisions on the evaluated workloads.
Figure~\ref{fig:predictor-tau}(a) shows that the online predictor overhead remains small (1.9\%--8.2\%) and decreases with larger batches and longer sequences, indicating that it is effectively amortized and unlikely to become a system bottleneck.


\textbf{Effect of $\tau$.} Figure~\ref{fig:predictor-tau}(b) shows the accuracy--TPOT tradeoff under different error-tolerance thresholds $\tau$. Increasing $\tau$ reduces TPOT at the cost of accuracy, with diminishing TPOT gains beyond 0.10 and more noticeable accuracy degradation at larger values. Overall, $\tau=0.10$ provides a favorable tradeoff.

\section{Related Work}
\label{relatedwork}

\textbf{Sparse Attention.} Sparse attention methods can be broadly categorized into static and dynamic sparsity. Static methods use predefined, input-independent patterns, such as Longformer~\cite{beltagy2020longformer}, StreamingLLM~\cite{xiao2023streamingllm}, and BigBird~\cite{zaheer2020big}, offering high efficiency but \wq{limited adaptivity to input-dependent attention patterns}. Dynamic sparse attention instead identifies important KV entries based on the input and can be further categorized by selection granularity into token-level and block-level methods. At token granularity, MagicPIG~\cite{chen2024magicpig} uses LSH to sample KV tokens, while PQCache~\cite{zhang2025pqcache} employs product quantization for approximate retrieval, avoiding exact top-$k$ selection over the entire KV cache. \wq{Such approximate retrieval methods reduce selection overhead, but introduce additional approximation and indexing/storage overhead.}

Block-level methods select groups of KV entries, amortizing selection and attention overhead across multiple KV entries. MInference~\cite{jiang2024minference}, SeerAttention~\cite{gao2024seerattention}, FlexPrefill~\cite{lai2025flexprefill}, and
XAttention~\cite{xu2025xattention} dynamically identify sparse patterns or important KV blocks, \wq{primarily targeting the compute-intensive prefill stage.} For decoding, Quest~\cite{tang2024quest}, InfLLM~\cite{xiao2024infllm}, Arkvale~\cite{chen2024arkvale}, and SeerAttention-R~\cite{gao2025seerattention} dynamically retrieve important KV blocks or distant context, while LServe~\cite{yang2025lserve} combines streaming heads with dynamic KV-page
selection. \wq{However, efficient CPU--GPU coordinated execution of sparse attention with CPU-resident KV remains less explored.}
Orthogonal to selection granularity, SparQ~\cite{SparQ2024}, ShadowKV~\cite{sun2024shadowkv}, and InfiniGen~\cite{lee2024infinigen} use compact or lower- dimensional representations to reduce the computation and/or data movement of Q--K matching. \wq{These techniques are complementary to token- and
block-level sparse selection.}

\textbf{KV Budget Allocation.} Existing budget allocation methods either assign retrieval budgets
using predefined rules, as in PyramidKV~\cite{cai2024pyramidkv}, or dynamically adapt the budget based on attention-related signals, such as attention scores or their distributions, as in MagicPIG~\cite{chen2024magicpig}, Progressive~\cite{zhou2025progressive}, and InfLLM~\cite{xiao2024infllm}.
\wq{However, these signals do not directly reflect the impact of KV sparsification on the attention output, potentially allocating unnecessary KV budgets to heads with limited output contribution.}

\textbf{Attention Parallelism.} Prior systems exploit CPU--GPU parallelism at different granularities. \wq{TwinPilots~\cite{yu2024twinpilots} pipelines CPU and GPU execution to overlap heterogeneous computation, while MoE-Lightning~\cite{cao2025moe} coordinates CPU--GPU execution for
MoE inference. NEO~\cite{jiang2025neo} offloads part of decoding attention to the CPU and overlaps it with GPU computation through load-aware scheduling.} These systems primarily coordinate
heterogeneous execution at the request, sub-batch, or operator level. A separate line of work parallelizes attention computation across GPUs. \wq{Megatron-LM~\citeN{shoeybi2019megatron} partitions Transformer
computation using model parallelism, while Ring Attention~\citeN{liu2023ring} and DeepSpeed Ulysses~\citeN{jacobs2023deepspeed} distribute long-sequence attention across devices using sequence/context parallelism.} These approaches primarily target GPU-side execution rather than sparse decoding over CPU-resident KV. In contrast, \sys{} dynamically schedules fine-grained sparse-attention tasks across CPUs and GPUs according to their estimated costs, exploiting both head-wise sparsity and heterogeneous device resources.

\section{Conclusion}
\label{conclusion}

This paper presents \sys{}, a CPU--GPU coordinated block-sparse attention system for efficient long-context decoding with CPU-resident KV. \sys{} combines output-contribution-aware budgeting, granularity-aware sparse configuration, and priority-based CPU--GPU scheduling to preserve near-FULL quality while improving decoding efficiency. In our evaluation, \sys{} improves TPOT by up to 3.7$\times$ over the best-quality fixed Quest configuration and achieves a better quality--efficiency tradeoff than StreamingLLM and MagicPIG. 
\wq{For models with model-native sparse attention, the built-in indexer
already performs fine-grained token selection, so \sys{}'s current output-aware budgeting and granularity selection are not directly applicable. Its CPU--GPU scheduling, however, is agnostic to the sparse-KV selection mechanism and can also be applied to model-native sparse workloads. \sys{} also remains applicable to full-attention components in hybrid attention architectures where sparsity is introduced at inference time.}


\begin{acks}
We sincerely thank the anonymous reviewers and our shepherd for their insightful feedback and constructive suggestions, which helped us improve the quality of this paper. This work was supported by the National Natural Science Foundation of China (NSFC) (62101012 and 62372436), the Beijing Natural Science Foundation (4262021), and the New Generation Artificial Intelligence National Science and Technology Major Project (2026ZD0128201).
\end{acks}

\clearpage
\bibliographystyle{ACM-Reference-Format}
\bibliography{main}


\end{document}